\documentclass[acmsmall,nonacm,screen]{acmart}
\usepackage{booktabs}
\usepackage{colortbl}
\usepackage{xspace}
\newcommand{\nul}{\textsc{null}\xspace}
\definecolor{hl}{RGB}{232,240,252}
\newcommand{\best}[1]{\textbf{#1}}
\newif\ifdraft \draftfalse

\title{What Egocentric Video Embeddings Encode and How to Combine Them with Inertial Sensors for Human Activity Recognition}
\author{Kansuke Habano}
\affiliation{\institution{Independent Researcher}\country{Japan}}

\authorsaddresses{}
\begin{document}
\begin{abstract}
Egocentric video and body-worn inertial sensors are increasingly recorded together,
and the video is usually encoded by pretrained models. What these embeddings
represent when the camera sits on the head, and how they should be combined
with inertial sensing, has received little attention. Across VideoMAEv2, CLIP,
DINOv2 and the egocentric VC-1, the leading principal directions of the embedding
are dominated by where the camera looks. Measured head orientation explains them on Ego-Exo4D for egocentric ($R^2$ $0.49$--$0.70$) but not exocentric cameras, and the scene layout explains them on
WEAR. The part that the limbs' inertial signals predict separates activities across
people. In a set-level probe, the rest separates exercise sets within a participant
but not activities across people. This is consistent with separate roles: the video finds the clips of the same set, and the
inertial classifier labels them. On WEAR, with temporal order removed and one random limb per one-second
clip, we propagate inertial predictions over a within-participant video graph whose
leading components are removed. This raises subject-wise cross-validated macro-F1 from $0.648$ to $0.852$.
In our comparisons, propagation helps only over the video graph, and fusing the modalities in one classifier, or adding inertial edges to the graph, performs worse.
\end{abstract}


\begin{teaserfigure}
\includegraphics[width=\textwidth]{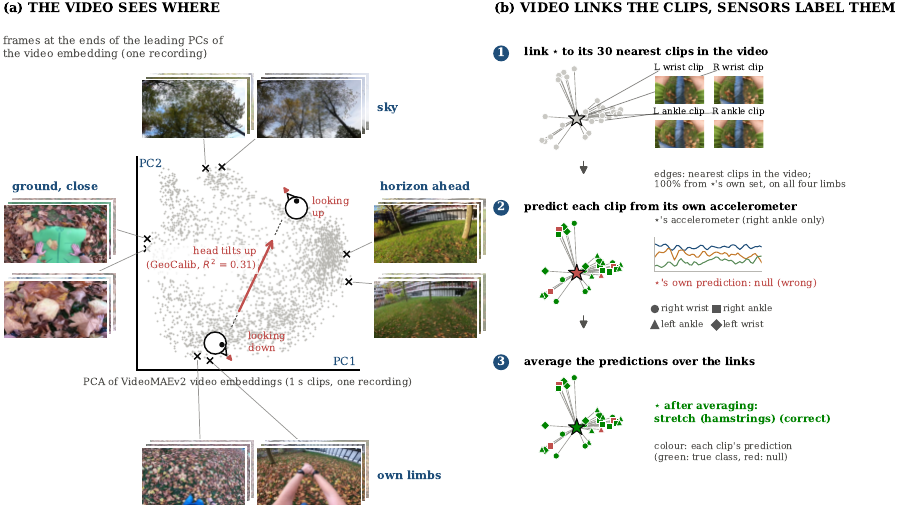}
\Description{Left: scatter of one recording's video embedding on its first two principal components, with camera frames at each end, each stacked on the frames 2 and 4 s later to show that it is video. The frames show close ground, the horizon ahead, the wearer's own limbs and the sky. An arrow shows the direction in which the estimated head tilt increases, with small head icons on its extension that look down at one end and up at the other. Right, three graphs stacked from top to bottom, each with the held-out clip as a star at the centre and edges to its 30 nearest clips in the video. (1) All nodes are grey, and photos of one neighbour per limb are linked to their nodes. (2) Nodes are coloured by each clip's own inertial prediction and shaped by its limb; the star is red (null), and the accelerometer trace of its right ankle is shown. (3) After averaging over the edges, the star is green, the true class.}
\caption{What egocentric video embeddings encode and how they combine with inertial sensing,
on one WEAR recording. (a)~Frames at the two ends of the first and second principal
components of the video embedding (VideoMAEv2). The components follow where the
camera looks. The arrow is the gradient of head tilt estimated from the frames
(GeoCalib~\cite{veicht2024geocalib}). Figure~\ref{fig:pcends} shows three participants. (b)~The video links the clips and the sensors label them, shown for a held-out clip $\star$ of a
training participant (cross-validation). (1)~$\star$ is linked to its 30 nearest clips in the video
(edges, with one photo per limb); all come from its own set and are spread over the four limbs.
(2)~Each clip is predicted from its own accelerometer (colour; shape: limb). As at test time,
$\star$ carries one limb, and its own prediction is wrong. (3)~Averaging the predictions over the
links corrects $\star$. Frames
from WEAR~\cite{bock2024wear} (CC BY-NC-SA 4.0, \url{https://creativecommons.org/licenses/by-nc-sa/4.0/}), downscaled.}
\label{fig:teaser}
\end{teaserfigure}

\maketitle

\section{Introduction}
\label{sec:intro}

Video from a head-worn camera is now routinely turned into features by pretrained
encoders. Generic encoders trained on web images and third-person video are applied
to it directly. Large egocentric corpora such as Ego4D~\cite{grauman2022ego4d}
have also produced encoders pretrained on first-person footage, such as
EgoVLP~\cite{lin2022egovlp}, LaViLa~\cite{zhao2023lavila} and, for embodied agents,
VC-1~\cite{majumdar2023vc1}. These encoders are built for egocentric tasks and
judged by their downstream accuracy. Little attention has been paid to what their
embeddings represent when the camera sits on the head, in particular how much of them
reflects where the head points rather than what the wearer does. The question matters
once such features are combined with body-worn inertial sensors. These are
increasingly recorded together with egocentric video, in datasets such as
WEAR~\cite{bock2024wear} and Ego-Exo4D~\cite{grauman2024egoexo4d}, and a head camera
rarely shows the wearer's body, only what the head points at. We analyse what
egocentric video embeddings encode and, from that analysis, propose how to use them
together with inertial sensing for human activity recognition (HAR).

Inertial-based HAR assigns an
activity label to each sliding window of an accelerometer stream. An important use case is the offline annotation of recorded wearable data. Annotation is
a documented bottleneck of the field. Labelling the Opportunity footage took 14--20 times
its duration~\cite{gunthermann2025annotators}, and the trade-off between label completeness
and accuracy is studied in its own right~\cite{hoelzemann2024annotation}. Vision foundation models applied to the
accompanying egocentric video have been used to reduce this effort~\cite{bock2024weak}. In this offline setting the recording is complete before inference begins, so all clips
of a participant can be processed together. Video is often available at collection time even when the
deployed sensor is inertial, and the placement of the device is not always under
the experimenter's control.

\begin{figure*}[t]
\centering
\includegraphics[width=\textwidth]{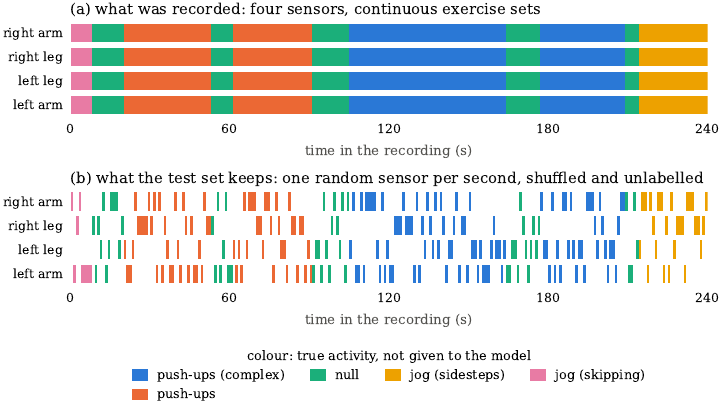}
\Description{Two timelines of one recording: the continuous activity sets of four sensors, and the single randomly assigned sensor per second that the test set retains.}
\caption{The setting on real data (240\,s of participant 0, coloured by the true
activity). (a) What was recorded: all four limb sensors and activity sets lasting
tens of seconds. (b) What the test set retains: each second retains \emph{one}
randomly assigned sensor, so the clips of one set come from different limbs. The
clips are given shuffled and without labels. A set still leaves many clips with
similar video, and the affinity graph exploits these.}
\label{fig:setting}
\end{figure*}

Existing HAR methods, including those benchmarked on WEAR~\cite{bock2024wear},
classify clips in their temporal context. A recent protocol
on WEAR~\cite{wear2026challenge} removes that context. Its test set is a
shuffled bag of one-second clips from four unseen participants. Each clip
carries the accelerometer of exactly \emph{one} randomly assigned limb
(Figure~\ref{fig:setting}) together with egocentric video features, and the stated
aim is ``to explore how to most effectively combine egocentric cameras with inertial
sensors''. Two obstacles
follow. First, the per-clip inertial evidence is thin, because a single limb over one second
often does not move during an exercise that moves another limb. Second, the
temporal context that would normally resolve such clips is gone, and the order
cannot be recovered from the video (Section~\ref{sec:setting}).

We start from the observation that the protocol withholds order but not
neighbourhood. Activity sets last tens of seconds, so each contributes many test
clips spread across all four limb placements. Pooling the per-clip
predictions within the true sets would raise macro-F1 from $0.648$ to $0.933$
without changing the classifier. Set membership is unavailable, but it can be read off from how the data arise. A set fixes both the activity and the circumstances in
which it is performed, namely the place, the posture, the direction the participant faces
and the scene in view. The activity drives the motion of the limbs and, through the
body, part of what the head camera sees. The circumstances drive most of the rest of
the view and are shared by every clip of the set, whichever limb carries its
sensor. We use this relation in reverse (Figure~\ref{fig:teaser}b). The video identifies clips
recorded under the same circumstances, that is, clips of the same set, and the inertial
classifier labels them. Propagating the
inertial predictions over a video graph then pools the four limbs over which a set's
clips are spread. We remove the leading directions of the video embedding, which mix the view
with the appearance of the activity, before building the graph. We use no test labels and
no temporal order (Section~\ref{sec:method}).

Our contributions are three-fold:
\begin{enumerate}\itemsep1pt
\item \emph{What egocentric video embeddings encode, and what degrades a graph built on
  them.} The leading principal directions of pretrained encoders applied to
  head-camera footage are dominated by where the camera looks. On
  Ego-Exo4D~\cite{grauman2024egoexo4d} they follow head orientation for four encoders, including the egocentric
  VC-1, but not for exocentric cameras. On WEAR they follow the scene layout. Removing them
  improves a within-participant graph for every encoder. Our analyses are consistent
  with an account in which the graph benefits from the view, which identifies the set,
  whereas the limb-predictable part in the same subspace degrades it
  (Sections~\ref{sec:encode} and~\ref{sec:removal}).
\item \emph{Evidence that, in the configurations we compared, building the graph from the video and starting propagation from inertial predictions works better than the fusion variants we tried.} In a controlled comparison with the same inertial classifier, the graph source sets the sign of the
  effect ($-0.060$ on an inertial graph, $+0.167$ on the video graph). A published test-time procedure~\cite{boudiaf2022lame} changes macro-F1 by
  $+0.001$ on the classifier's own features and by $+0.118$ on the video graph. Every
  fusion variant we tried performs worse, although none was tuned as extensively as
  the proposed system (Sections~\ref{sec:roles} and~\ref{sec:stages}).
\item \emph{A simple instantiation and what it shows about the evaluation protocol.}
  Built from standard components, the method raises subject-wise cross-validated macro-F1 from
  $0.648$ to $0.852$ without temporal order or test labels. We also report two
  properties of the protocol rather than of the method
  (Sections~\ref{sec:method}, \ref{sec:main} and~\ref{sec:discussion}).
\end{enumerate}

The remainder of the paper is organised as follows. Section~\ref{sec:related}
reviews related work, Section~\ref{sec:encode} analyses what egocentric video
embeddings encode, and Section~\ref{sec:setting} defines the recognition setting.
Section~\ref{sec:method} describes the approach built on that analysis and Section~\ref{sec:setup} the
experimental setup. Section~\ref{sec:results} reports results, and
Section~\ref{sec:discussion} discusses their meaning and implications for the
evaluation protocol. Sections~\ref{sec:limitations} and~\ref{sec:conclusion} state
limitations and conclude. Additional analyses, including the complete list of
tested explanations, are in the appendix.

\section{Related Work}
\label{sec:related}

\subsection{Representations of egocentric video}
Egocentric video was first encoded with models pretrained on web images and
third-person video, and the strongest video--language models relied on third-person
corpora. Large egocentric corpora such as Ego4D~\cite{grauman2022ego4d} made
pretraining on first-person footage possible. EgoVLP~\cite{lin2022egovlp} adapts
video--language contrastive learning to egocentric clips, LaViLa~\cite{zhao2023lavila}
trains on narrations that a language model generates for them, and
VC-1~\cite{majumdar2023vc1} learns visual representations for embodied agents by
masked auto-encoding on egocentric video. These encoders target egocentric tasks such
as video--text retrieval, action recognition, episodic memory and robot control, and
are evaluated by downstream accuracy. Generic encoders such as VideoMAEv2, CLIP and
DINOv2 remain in common use. Head motion, a defining trait of such footage, has been
exploited as a cue, for example by EgoDistill~\cite{tan2023egodistill}, which reconstructs video features
cheaply from head IMU signals. Probing studies ask what pretrained embeddings encode,
for example properties of 3D scenes~\cite{man2024lexicon3d}. For word embeddings,
the leading principal directions are known to dominate similarities, and removing them
improves the representation~\cite{mu2018allbut}, in relation to hubness~\cite{radovanovic2010hubs}. We
know of no work that relates the leading directions of egocentric video embeddings to
the orientation of the head, or asks how such embeddings should be used when they
are combined with body-worn sensors.

\subsection{Inertial HAR and the WEAR benchmark}
WEAR~\cite{bock2024wear} pairs four limb-worn accelerometers with a head-mounted
camera. It benchmarks inertial, video and fusion models for activity recognition
of its eighteen workout activities and for temporal action localisation, under
leave-one-subject-out (LOSO) cross-validation. Temporal action localisation has since
been shown to apply to inertial data on its own~\cite{bock2024tal}.
The two earlier editions of the challenge were inertial-only. Van Der Donckt et
al.~\cite{vanderdonckt2024swap} used all four limb sensors with temporal context
and reached $0.90$--$0.92$ macro-F1, exploiting left--right symmetry and
upper--lower limb pairing. The second edition restricted inference to a single
sensor. Phan et al.~\cite{phan2025robust} combined hand-crafted features with
sensor-rotation augmentation and a gradient-boosting ensemble ($0.588$
macro-F1), and other teams addressed the dominance of the \nul
class~\cite{harma2025null} and used reservoir computing~\cite{vedanam2025reservoir}.
Test splits differ between editions, so these numbers serve to locate results and do not allow a direct comparison.
Inertial features are sensitive to where the device is worn. Existing remedies make
the classifier insensitive through orientation-invariant features, symmetric limb
pairing~\cite{vanderdonckt2024swap} or large multi-placement
corpora~\cite{adaimi2024location}. Training with four sensors and testing with one
is an instance of learning under privileged information~\cite{lopezpaz2016unifying}.

\subsection{Egocentric video and inertial sensing}
Recent work couples egocentric video and inertial data to improve or compress the
\emph{classifier}. COMODO~\cite{chen2025comodo} distils a frozen video encoder
into an IMU encoder so that inference can run on the IMU only. These are
cross-modal transfers at training time in an inductive setting. Closest to our use of
the video, Bock et al.~\cite{bock2024weak} embed the egocentric video of WEAR and two
other datasets with vision foundation models such as CLIP and cluster the embeddings.
An annotator labels the centroid clip of each cluster, and this label is propagated to
the corresponding inertial data. We share the premise that egocentric embeddings
group clips of the same activity. However, we propagate the predictions of an inertial
classifier instead of human labels, at test time within each unseen participant, and we
ask which part of the embedding should be removed before grouping. We also assign the
modalities different roles at inference time. In our setting a privileged
four-sensor teacher brings no improvement once the transductive structure is in place
(Section~\ref{sec:roles}). The video features we use are frozen
VideoMAEv2~\cite{wang2023videomae} embeddings distributed with the dataset.

\subsection{Transduction and test-time adaptation}
Adapting to an unseen person with unlabelled test data is an established setting
for wearable HAR. Wang et al.~\cite{wang2024oftta} adapt without back-propagation,
and SIGHT~\cite{zhou2026sight} exploits the temporal structure between consecutive
clips, which our protocol removes. Label propagation on a $k$-NN graph
is classical~\cite{zhou2003local} and remains competitive with deep
features~\cite{iscen2019label}. Transductive few-shot methods such as
TPN~\cite{liu2019tpn} learn the propagation itself, and
LAME~\cite{boudiaf2022lame} adjusts a black-box classifier's outputs at test time
with a $k$-NN Laplacian and no labels. In these methods the affinity graph is built
from the same representation as the classifier, and in label propagation the graph
contains labelled nodes. Our graph has no labelled nodes. It smooths
predictions among unlabelled test clips and is built from a different
modality. Mutual $k$-NN graphs follow the $k$-reciprocal idea of re-ranking in
retrieval~\cite{zhong2017reranking}.

\subsection{Smoothing predictions over another source's neighbourhoods}
Smoothing a classifier's per-sample posteriors over a neighbourhood structure
derived from a different source has a long history outside HAR. In hyperspectral
remote sensing, pixelwise SVM probabilities from the spectrum are regularised by a
Markov random field over spatial position~\cite{tarabalka2010svm} or by
edge-preserving filtering guided by the image's leading principal
components~\cite{kang2014edge}. The dense CRF~\cite{krahenbuhl2011dense} and the
bilateral solver~\cite{barron2016bilateral} smooth network outputs with affinities
from a reference image. In graph learning,
Correct-and-Smooth~\cite{huang2021cs} and APPNP~\cite{klicpera2019appnp} propagate
the predictions of a structure-agnostic base model over a given graph, and
population-graph GCNs~\cite{parisot2018population} assign edges and node features
to different sources by design.  When
several views are available, graph-based learning usually \emph{fuses} them. Examples are
sparse integration of view Laplacians~\cite{karasuyama2013sparse},
auto-weighting by the smoothness of the current solution under each
view~\cite{nie2016autoweighted} and per-sample modality weights as in the weighted
nearest neighbours of multimodal single-cell analysis~\cite{hao2021wnn}. Our
contribution is a controlled comparison in a wearable setting. It shows that fusion
of either kind performs worse than a division of roles in which the video builds the graph and
inertial information enters it only through predictions.
\section{What Egocentric Video Embeddings Encode}
\label{sec:encode}

In this section we ask what a pretrained encoder represents when it is applied to
head-camera footage. The egocentric pretraining work of Section~\ref{sec:related}
leaves this question open. The question does not depend on the recognition task, so
we answer it on the recordings themselves, before we define the task in
Section~\ref{sec:setting}.

\subsection{Data and encoders}
\label{sec:enc-data}
Table~\ref{tab:data} summarises the data. WEAR~\cite{bock2024wear} has 22 training
participants and 24 recordings of 18 outdoor workout activities. It provides 3-axis
accelerometers at 50\,Hz on both wrists and both ankles and 768-dimensional
VideoMAEv2 frame embeddings at 30\,fps from a head-mounted camera. Each exercise is
performed in sets, which are continuous bouts at one place with a median length of
$32$\,s. Time not annotated as an activity, such as breaks between sets or walking
to the next station, is labelled \nul. We cut each recording into one-second clips.
A clip holds the video features and the accelerometer signal of that second, and it
is the unit that is classified. In this section we analyse the clips of the training
participants.

\begin{table}[t]
\centering
\caption{Datasets. WEAR is used for the method and every evaluation. Ego-Exo4D is used
only for the analysis of Section~\ref{sec:encode}.}
\label{tab:data}
\footnotesize
\setlength{\tabcolsep}{3pt}
\begin{tabular}{l>{\raggedright\arraybackslash}p{3.3cm}>{\raggedright\arraybackslash}p{2.7cm}>{\raggedright\arraybackslash}p{3.4cm}}
\toprule
 & WEAR, training & WEAR, test & Ego-Exo4D (analysis only) \\
\midrule
Participants & 22 (24 recordings) & 4, unseen & 6 (79 takes) \\
Activities & 18 outdoor workouts and \nul & same & basketball, dance, soccer \\
Inertial & 3-axis acc.\ at 50\,Hz on both wrists and ankles & one randomly assigned limb per clip & head trajectory (6 DoF) of the Aria glasses \\
Video & head camera: VideoMAEv2 embeddings; public raw video & VideoMAEv2 embeddings & egocentric (Aria) and one exocentric camera \\
Clips & about $69{,}000$ one-second tiles, in order & $12{,}234$, shuffled & $8{,}163$ one-second clips \\
Labels & per second & hidden & not used \\
\bottomrule
\end{tabular}
\end{table}

Besides the VideoMAEv2 embeddings of the dataset, we use the CLIP
ViT-B/32~\cite{radford2021clip} and DINOv2 ViT-B/14~\cite{oquab2024dinov2} image
encoders and VC-1 ViT-B/16~\cite{majumdar2023vc1}, which is pretrained by masked
auto-encoding on over 4{,}000 hours of egocentric video and ImageNet. We apply these
encoders to the public WEAR videos of the training participants
(Appendix~\ref{app:impl}). WEAR has no head sensor. To relate the embedding to the
head, we use Ego-Exo4D~\cite{grauman2024egoexo4d}, whose head-mounted Aria glasses
record a six-degree-of-freedom head trajectory. We take two participants from each of three whole-body domains (basketball, dance, soccer; 79 takes, $8{,}163$ one-second
clips). We encode their egocentric video with the same four encoders and the
extraction specification of the WEAR features. Per clip, we compute the head tilt
(gravity direction in the device frame), head height, and rotation and translation
speed. For comparison we encode the first exocentric camera of the same takes in the
same way. We use Ego-Exo4D for this analysis only, and no part of the method is
trained or tuned on it.

\subsection{The leading components track head orientation on Ego-Exo4D}
\label{sec:enc-ego}
A head camera never shows the performer's body as a whole. The frame contains what
the head points at, such as the ground, trees and sky, buildings, and the wearer's
hands and feet. On Ego-Exo4D the head trajectory is measured. There, head
orientation (tilt, height and heading) predicts the first two components of the
egocentric embedding within a take with $R^2$ $0.49$--$0.70$ across the four
encoders. The egocentric VC-1 gives the highest value (Table~\ref{tab:encoders}).
Head tilt is linearly decodable from the top-30 subspace, better than from 30 random directions ($0.74$--$0.91$ against $0.61$--$0.79$). Tilt is thus
readable throughout the embedding but most strongly from the leading subspace. None of this holds for the exocentric cameras of the same takes encoded
in the same way. For them, head orientation predicts the first two components with
$R^2 \le 0.20$, and tilt is only slightly more decodable from the top-30 subspace
than from 30 random directions. The exocentric camera sees the same movements from a
fixed position, so this control separates what the viewpoint contributes from what the
activity itself contributes. All four encoders concentrate the view in the leading
directions on egocentric footage, and none does so on the exocentric footage of the
same takes. This concentration therefore appears to be a property of the
head-mounted viewpoint rather than of a particular encoder.

\begin{figure}[t]
\centering
\includegraphics[width=\textwidth]{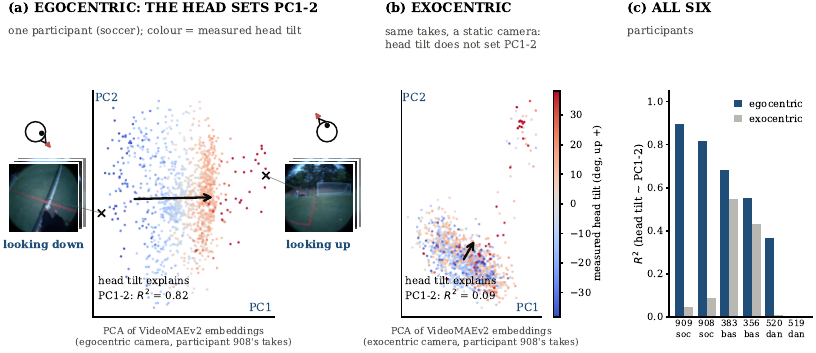}
\Description{Left: scatter of one participant's egocentric video embeddings on their first two principal components, coloured by measured head tilt. An arrow marks the direction in which the tilt increases, with egocentric frames at its two ends (looking down at the ground, looking up across the field) and small head icons. Middle: the same takes seen by the exocentric camera, where the colours are mixed and the arrow is short. Right: bars of the variance of head tilt explained by the two components for six participants. The bars are higher for the egocentric camera.}
\caption{What the leading components of an egocentric embedding follow, on Ego-Exo4D
(VideoMAEv2, centred within each take). (a)~Egocentric camera of one participant
(soccer), coloured by the measured head tilt. The components follow the tilt. The
arrow is the direction in which the tilt increases, with length proportional to the
correlation, and the frames are the clips at its two ends. (b)~Exocentric camera of
the same takes. Here the components do not follow the tilt. (c)~$R^2$ of head tilt
from components 1--2 for all six participants. Frames from
Ego-Exo4D~\cite{grauman2024egoexo4d}, used under the Ego-Exo4D licence agreement,
downscaled.}
\label{fig:egoexo}
\end{figure}

Figure~\ref{fig:egoexo} shows this for one participant. When coloured by measured
head tilt, the egocentric embedding is ordered from looking down to looking up along
its first component. The exocentric embedding of the same takes shows no such order.
Within takes, head tilt explains components 1--2 of the egocentric camera with $R^2$
between $0.00$ and $0.90$ across the six participants (median $0.62$; near zero for one
dance participant). These values
are in-sample, unlike the cross-validated values of Table~\ref{tab:encoders}. For
the exocentric camera the value is near zero except in basketball ($0.43$ and
$0.55$), where the head moves with a visible body motion such as a jump shot. The
direction that encodes head tilt is shared across people. A linear read-out trained
on five participants predicts the sixth participant's tilt with correlation $0.87$--$0.97$ for VideoMAEv2 ($0.79$--$0.98$ across the four encoders).

\begin{table}[t]
\centering
\caption{The leading components of egocentric and exocentric embeddings. Layout and
Orient.: within-recording $R^2$ of predicting components 1--2 from the scene layout
of the frame or from head tilt, height and heading (cross-validated over five time
blocks). Tilt$\leftarrow$top-30: $R^2$ of decoding head tilt from the top-30
subspace (cross-validated over takes), with 30 random directions in brackets. WEAR
has no head sensor.}
\label{tab:encoders}
\setlength{\tabcolsep}{2.5pt}
\begin{tabular}{llccc}
\toprule
Encoder & Data, view & Layout & Orient. & Tilt$\leftarrow$top-30 \\
\midrule
VideoMAEv2 & WEAR, ego & 0.52 & --- & --- \\
           & Ego-Exo4D, ego & 0.60 & 0.49 & 0.74 (0.63) \\
           & Ego-Exo4D, exo & $-0.23$ & 0.07 & 0.36 (0.27) \\
CLIP       & WEAR, ego & 0.61 & --- & --- \\
           & Ego-Exo4D, ego & 0.64 & 0.51 & 0.78 (0.61) \\
           & Ego-Exo4D, exo & 0.44 & 0.05 & 0.22 (0.20) \\
DINOv2     & WEAR, ego & 0.58 & --- & --- \\
           & Ego-Exo4D, ego & 0.72 & 0.58 & 0.84 (0.71) \\
           & Ego-Exo4D, exo & 0.46 & 0.20 & 0.33 (0.25) \\
VC-1       & WEAR, ego & 0.53 & --- & --- \\
           & Ego-Exo4D, ego & 0.81 & 0.70 & 0.91 (0.79) \\
           & Ego-Exo4D, exo & 0.54 & $-0.14$ & 0.20 (0.16) \\
\bottomrule
\end{tabular}
\end{table}

\subsection{Scene layout and head tilt on WEAR}
\label{sec:enc-wear}
WEAR has no head sensor, so we measure what the view contains. We take the spatial
layout of scene classes from a semantic segmentation of each clip's frame
(Appendix~\ref{app:impl}). Within a recording, this layout predicts components 1--2
with $R^2$ $0.52$--$0.61$, about as well as for the egocentric Ego-Exo4D video
($0.60$--$0.81$). The wearer's visible body parts reach $0.05$--$0.08$, and nothing
predicts components beyond the tenth. The layout itself is not specific to the
egocentric view, because a person moving in a static exocentric frame also changes it
($0.44$--$0.54$ for CLIP, DINOv2 and VC-1. For VideoMAEv2 the cross-validated $R^2$ is
$-0.23$, below that of predicting the mean). The specific part is that in the
egocentric view the layout changes because the head moves. We train the tilt
read-out on all of Ego-Exo4D and apply it to WEAR, which was recorded with a
different head camera. Its output agrees with the per-frame camera pitch estimated
there by GeoCalib~\cite{veicht2024geocalib} (median correlation per recording
$0.58$, $0.70$ and $0.79$ for VideoMAEv2, CLIP and DINOv2). It also ranks lying
exercises as the most upward-looking. However, tilt accounts for only part of
components 1--2 on WEAR (median $R^2$ $0.10$--$0.34$). There, the leading components
depend as much on where the wearer is and which way they face.

\subsection{The view marks the set rather than the activity}
\label{sec:enc-set}
We test this division as follows. We split each embedding into its limb-predicted
component, predicted out of fold from the four limbs' inertial features, and the
rest. We then average both over each set (608 activity sets of at least ten clips).
Below we refer to the two parts as the appearance of the activity and the
circumstances. These names are interpretations. The split is defined by what the four limbs
predict, it does not identify independent factors, and what four limbs predict need not be what a
single-limb classifier already uses.
Between sets of different participants, whose place, day and person differ, the
limb-predicted component still separates the activities. The probability that a pair
of different activities lies farther apart than a pair of the same activity is
$0.91$ (95\% interval over participants $[0.88, 0.93]$). For the rest, the
probe detects no activity separation ($0.52$, $[0.47, 0.58]$). It compares set averages by
cosine distance, so it does not exclude activity information that a non-linear read-out
or another metric could extract. Within a participant the rest separates
sets instead. Two halves of one set lie $0.08$ apart in cosine distance, as do
adjacent sets of the same exercise. Adjacent sets of different exercises lie $0.93$
apart, nearly as far as sets of another participant ($0.99$), whatever the gap
between them. A new exercise comes with a new posture and facing, so the
circumstances belong to the set. In this probe they do not indicate which activity it is. CLIP,
DINOv2 and VC-1 give the same pattern. By construction the rest contains what the
limbs do not predict, so we expect it to show little activity separation across people.
The informative result is that it still separates sets within a participant.

\subsection{Implications for combining the modalities}
\label{sec:enc-impl}
The leading directions of an egocentric embedding thus hold the view of the moment
(where the head points, in which place and posture). Within that view they hold the
appearance of the activity. The clips of a set share the view, and it changes from
set to set. Linking clips with a similar view therefore finds the clips of the same
set. In our probe the view shows no activity separation that transfers across people, so we do not rely on it to identify what they are doing. The appearance does encode the activity, but it is predictable from the limbs'
inertial signals, and propagation pools the inertial predictions of all four limbs
within a set (Section~\ref{sec:roles}). Seen from the head, an exercise also shares
its appearance with the pauses around the set. These observations suggest a division of roles. The
video builds the graph, the inertial classifier supplies the labels, and we remove
the appearance from the graph. Section~\ref{sec:method} builds a method on this
division. Section~\ref{sec:removal} tests which part of the embedding the graph is
better without, and Section~\ref{sec:roles} tests the division against fusion.

\section{Problem Setting}
\label{sec:setting}

\paragraph{Test protocol.}
We evaluate the method under a protocol on WEAR~\cite{wear2026challenge} that
removes temporal order (Figure~\ref{fig:setting}). The test set contains
$12{,}234$ one-second clips (windows in the challenge data) from four unseen
participants. In the training recordings, \nul{} covers $40\%$ of the clips (from $14\%$ to $57\%$ per recording). Each clip provides a $(50,3)$
accelerometer segment from one limb with known placement. It also provides 15 rows of
VideoMAEv2 embeddings, one per frame, each computed from the 16 frames around that
frame (Appendix~\ref{app:impl}). The clips are consecutive, non-overlapping
one-second tiles of each recording, shuffled. Each second of a recording appears once
with one randomly assigned limb.

\paragraph{Task and metric.}
The task is 19-way classification (18 activities and \nul) of each clip. It is
evaluated by the macro F1-score over all 19 classes, including the \nul{} class.
Segment-based metrics need start and end times and are undefined for a shuffled bag
of clips. We therefore report record-based precision, recall and F1-score.

\paragraph{What the protocol removes.}
Temporal order cannot be restored from the video. We tried nearest-neighbour matching
of clip-mean video features, directional matching from the last frame of one clip to
the first frame of the next, and a global linear assignment with unique successors.
All of them plateau at successor top-1 accuracy $\le 0.35$, which is too low for
temporal models. Two properties of the data survive. Activity sets have median length
$32$\,s, so each set contributes many clips spread over all four placements. In
addition, the top-10 video neighbours of a clip share its label 77--88\% of the time.
The raw accelerometer signal does reveal adjacency for the quarter of consecutive
pairs that happen to share a limb. We do not use it, and we discuss it in
Section~\ref{sec:disc-bench}.

\paragraph{Transductive setting.}
Recordings are complete before inference, so we treat each test participant's clips
jointly. All video features and model outputs of that participant may be used at test
time, but test labels may not. Every model is trained on training participants only.

\section{Method}
\label{sec:method}

\begin{figure*}[t]
\centering
\includegraphics[width=\textwidth]{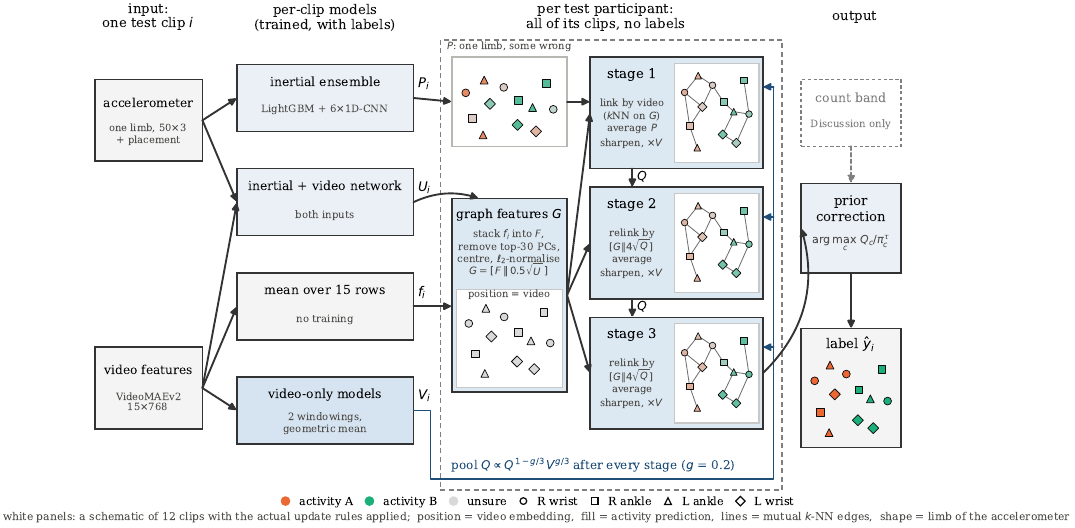}
\Description{Block diagram of the method. Per-clip inertial and video classifiers feed a per-participant video graph. Predictions are propagated on this graph over three stages with graph re-estimation, followed by prior correction.}
\caption{Overview of the approach, from one test clip to its label. Left: the two
inputs of a clip. Centre-left: per-clip models trained on the training
participants. $P$, $U$ and $V$ are 19-class probabilities and $f$ is the mean of the
clip's 15 video rows. Centre-right: everything inside the dashed box runs once
per test participant on all of that participant's clips and uses no labels. This
includes the graph features $G$, three propagation stages with the graph rebuilt from
$[G \,\|\, 4\sqrt{Q}]$ between stages, and the video-only probabilities $V$ pooled
after every stage. Right: the decision rule yields the label. The white panels are
a schematic of a dozen clips with the actual update rules applied (position =
video embedding, fill = activity prediction, grey = unsure, lines = mutual $k$-NN
edges, shape = limb carrying the accelerometer). The inertial prediction $P$ has
wrong and unsure clips. Stage~1 links clips that look alike in video and averages their
predictions, and stages~2--3 relink using the refined predictions as well. Raw inertial features never enter the graph. Inertial information
enters it only through class probabilities ($U$ in $G$ and the refined $Q$), and the
video never enters the inertial classifier. The dashed count band is the protocol
constraint of Section~\ref{sec:disc-bench} and is not part of the reported system.}
\label{fig:pipeline}
\end{figure*}

The method implements the division of roles suggested by Section~\ref{sec:encode}.
The video builds a graph that links clips showing the same scene, that is, clips of
the same set. The inertial classifier supplies the labels, and we remove the
appearance of the activity from the graph. Figure~\ref{fig:pipeline} gives an
overview. Per-clip classifiers produce class probabilities (Section~\ref{sec:m-base}).
For each test participant we build a graph from the video embeddings
(Section~\ref{sec:m-graph}). We propagate the inertial predictions over it in stages
that alternate with re-estimation of the graph (Section~\ref{sec:m-prop}). Video
evidence and a decision rule complete the pipeline (Section~\ref{sec:m-decision}).

\subsection{Per-clip classifiers}
\label{sec:m-base}
The inertial classifier produces $P_i \in \Delta^{19}$ for clip $i$ from its
single-limb accelerometer segment and placement. It is a weighted blend of two parts.
The first is LightGBM on 78 hand-crafted features (per-axis moments, quantiles,
zero-crossing rate, jerk statistics, orientation-invariant magnitude features,
inter-axis correlations, spectral band powers and entropy) with a placement
one-hot, at weight 1.5. The second is six 1D-CNNs on the raw segment with a placement
embedding, at weight 1.0 each. Three of them use rotation augmentation
($\pm20^\circ$ to $\pm30^\circ$), and three additionally take the 78 features as an
auxiliary branch. Each training clip is expanded into four samples, one per
placement. Two further classifiers use the video. A network on both modalities gives
probabilities $U_i$ that enter the graph features. A video-only network gives
probabilities $V_i$ that are pooled after propagation. Architectural details are in
Appendix~\ref{app:impl}.

\subsection{Participant-level video graph}
\label{sec:m-graph}
For each test participant, let $f_i \in \mathbb{R}^{768}$ be the mean of clip
$i$'s 15 video rows and $F$ the matrix of these vectors. We remove the top $m=30$
principal components of $F$, computed within the participant, then centre and
$\ell_2$-normalise the rows. The graph features are
\begin{equation}
G_i = \bigl[\, F_i \;\big\|\; \gamma \sqrt{U_i} \,\bigr], \qquad \gamma = 0.5,
\end{equation}
re-centred and row-normalised. We connect each clip to its $k=30$ nearest
neighbours by cosine similarity $s_{ij}$ and retain only mutual edges. A $k$-NN graph
needs neither the number of sets nor their boundaries, which are unknown, and $k$ is
close to the median set length of $32$ clips. Hard clustering did no better
(Appendix~\ref{app:negative}). The affinity
is $W_{ij} \propto \exp(10\, s_{ij})$ over retained edges, row-normalised. All
operations use only the participant's own clips, with no temporal order and no
labels. Section~\ref{sec:encode} motivates the component removal and
Section~\ref{sec:removal} analyses it.

\subsection{Propagation with graph re-estimation}
\label{sec:m-prop}
Let $P \in \mathbb{R}^{n\times 19}$ collect the participant's inertial predictions.
Each stage propagates
\begin{equation}
Q^{(t+1)} = (1-\alpha)\, Q^{(0)} + \alpha\, W Q^{(t)}, \qquad \alpha = 0.75,
\end{equation}
for five iterations from $Q^{(0)}$, which is $P$ in the first stage and the output
of the previous stage afterwards. It then sharpens the result,
$Q \leftarrow Q^{1.4}$ renormalised. Between stages the graph is rebuilt from
$[\, G \,\|\, \lambda \sqrt{Q}\,]$ with $\lambda = 4$, so that clips whose
refined predictions disagree are pushed apart. We use three stages. The stages
therefore alternate between estimating the labels given the graph and
re-estimating the graph given the labels. Section~\ref{sec:stages} tests this
reading.

\subsection{Video evidence and decision rule}
\label{sec:m-decision}
After every stage the video-only probabilities are pooled log-linearly,
\begin{equation}
Q \;\propto\; Q^{1-g/3}\, V^{g/3}, \qquad g = 0.2,
\label{eq:pool}
\end{equation}
so that the total video weight is $g$ over the three stages. Macro-F1 favours recall on rare classes, so the posterior mode is not the best decision. The final label is therefore
$\hat{y}_i = \arg\max_c Q_{ic} / \pi_c^{\tau}$, where $\pi_c$ is the mean predicted
probability of class $c$ over all clips being labelled (at test time all test clips,
in cross-validation the held-out clips of all participants) and $\tau$ is a single scalar selected by
cross-validation ($0.9$ for the final system). Averaging $\pi_c$ over each participant's clips instead
gives higher cross-validated scores (Section~\ref{sec:unseen}), but we report the rule that we
fixed before the hidden test and used there.

\section{Experimental Setup}
\label{sec:setup}

\paragraph{Evaluation protocol.}
We use five-fold cross-validation grouped by participant, with the two recordings
of one person in the same fold. The benchmark's leave-one-subject-out protocol is
its $k=n$ limit. We build the graph and propagate per held-out participant, as at
test time. The class prior of the decision rule is averaged over all clips being
labelled (Section~\ref{sec:m-decision}). The validation data mirror the test structure, with non-overlapping
one-second tiles that each have one uniformly random limb. Because that assignment is
random, we repeat it over seeds (three for the main results, two for the
comparisons unless stated) and report means. The seed-to-seed standard deviation
is about $0.002$, and we treat differences below $0.005$ as inconclusive. For the
main gains we also report 95\% intervals from resampling the 22 participants
(2{,}000 draws), which capture the variation between people. We selected all
hyperparameters on this cross-validation. The training epoch of each network and the
threshold $\tau$ of each configuration are also chosen on the held-out folds, with the
labels against which we report. The reported scores are therefore the cross-validated results of an
exploratory design process, not unbiased estimates of generalisation, and the
intervals do not include the uncertainty of model selection. The seeds repeat the limb
assignment, not the selection. The hidden test set of the challenge
(Section~\ref{sec:main}) is the only evaluation outside this loop.

\paragraph{Compared variants.}
Beyond component ablations, we compare against the following. (i) Fusion at the
classifier: a base that consumes both modalities, and a blend of the two bases.
(ii) Fusion at the graph: adding inertial $k$-NN edges, learning view weights by the
auto-weighting criterion of~\cite{nie2016autoweighted}, and per-clip weights in the
style of~\cite{hao2021wnn}. (iii) A different propagation rule,
LAME~\cite{boudiaf2022lame}, on either graph. (iv) A four-sensor inertial teacher,
used directly or by distillation. (v) A graph pooled across participants.
Models that need an ordered stream do not apply. Throughout, fusion means combining
the two modalities in one classifier or in the edges of the graph. The method itself
also uses both modalities: it adds the probabilities of video models to the graph features and pools them after
every stage (Section~\ref{sec:method}). The comparison is therefore not between one
modality and two, but between where the video enters. In the method it builds the graph
and enters the labels only as a weighted prediction, whereas in the variants it enters the classifier's input or the graph's edges. The
comparison is not matched in tuning effort. The inertial base is an ensemble of
LightGBM and six networks, the classifier on both modalities is a single network
with its own training schedule, and the variants reuse the propagation settings tuned
for the proposed system.

\paragraph{Use of AI tools.}
The code that implements the methods, experiments and analyses was written, and the
figures were produced, with an AI coding assistant (Claude Code, Anthropic) under the
author's direction. The author designed the study and the analyses, checked every reported number against the code to be released, and takes responsibility for all content.

\paragraph{Data and ethics.}
We use only public datasets (WEAR and Ego-Exo4D), which their authors collected with
the participants' consent and which we use under their licences. No new data were
collected. The figures show the scene and the wearer's own limbs. Test labels were never used,
all hyperparameters were selected on subject-wise cross-validation, and the transductive
steps use only test-set features and model outputs. Raw video was used for training
participants only.
\section{Results}
\label{sec:results}

We report the results in five parts. Part (1) gives the gain of each component of
the method (Section~\ref{sec:main}). Parts (2) and (3) test the analysis of
Section~\ref{sec:encode}. They ask what component removal takes out of the video
embedding (Section~\ref{sec:removal}) and whether the modalities should be fused
or given separate roles (Section~\ref{sec:roles}). Part (4) covers what the
propagation stages do (Section~\ref{sec:stages}). Part (5) covers robustness to
unseen participants, to clips that arrive in parts and to the hyperparameters
(Section~\ref{sec:unseen}).

\subsection{Main results}
\label{sec:main}

\begin{figure*}[t]
\centering
\includegraphics[width=\textwidth]{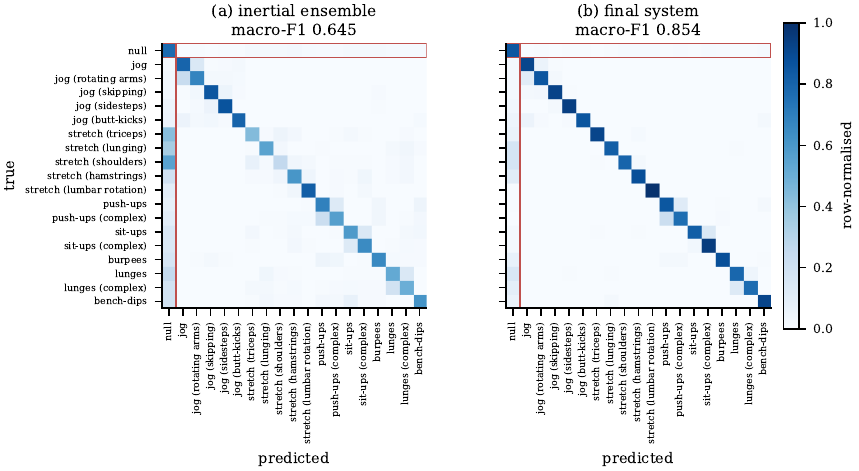}
\Description{Two row-normalised confusion matrices over 19 classes, for the inertial ensemble and for the final system. The final system is much closer to diagonal.}
\caption{Row-normalised confusion matrices before and after the transductive stages (seed 42). The \nul row and column are outlined. The inertial classifier's dominant
error is activity~$\rightarrow$~\nul, and propagation removes most of that column.}
\label{fig:confusion}
\end{figure*}

Table~\ref{tab:prf} reports record-based precision, recall and F1-score for the
successive components, split into the \nul{} class and the 18 activities. The
inertial classifier reaches $0.648$ macro-F1 on single clips. The earlier
single-sensor result on WEAR, $0.588$~\cite{phan2025robust}, was obtained on a
different test split. Propagation over the video graph raises macro-F1 by $+0.182$, as do pooling the
video-only probabilities ($+0.013$) and removing the leading components of the
video features ($+0.009$), which gives $0.852$.
When we resample participants, the 95\% intervals of the three gains are $[+0.167,
+0.196]$, $[+0.009, +0.017]$ and $[+0.005, +0.013]$. The three gains hold for 22, 19 and 18 of the 22 participants.
Recall increases more than precision, as the decision rule intends. The transductive stages benefit the activities far more than \nul, and the gap between
$F_1^{\mathrm{NULL}}$ and $F_1^{\mathrm{act}}$ shrinks from $0.119$ to $0.026$. The
confusion matrices (Figure~\ref{fig:confusion}) show where this happens. The
inertial classifier's dominant error is activity~$\rightarrow$~\nul, because a
limb at rest during an exercise that moves another limb is inertially
indistinguishable from a break. Propagation removes most of that column because
other clips of the same set, recorded on other limbs, outvote it.

\begin{table}[t]
\centering
\caption{Record-based results under subject-wise cross-validation (mean over three
seeds). $F_1$ is macro-F1 over all 19 classes. The last two columns split it into
the \nul class and the 18 activities.}
\label{tab:prf}
\setlength{\tabcolsep}{4pt}
\begin{tabular}{lccccc}
\toprule
 & P & R & $F_1$ & $F_1^{\mathrm{NULL}}$ & $F_1^{\mathrm{act}}$ \\
\midrule
Inertial classifier      & 0.658 & 0.644 & 0.648 & 0.761 & 0.642 \\
\; + propagation         & 0.825 & 0.840 & 0.830 & 0.861 & 0.828 \\
\; + video pooling       & 0.836 & 0.854 & 0.843 & 0.872 & 0.841 \\
\rowcolor{hl}
\; + component removal   & \best{0.842} & \best{0.865} & \best{0.852} & \best{0.876} & \best{0.850} \\
\bottomrule
\end{tabular}
\end{table}

Pooling the inertial predictions within the true activity sets gives an upper
bound of $0.933$. The approach closes $72\%$ of the gap between the inertial
classifier and this bound. Per class (Appendix~\ref{app:perclass}), the largest
gains are on classes that a single placement cannot capture. A leg sensor all but misses
arm stretches (recall $0.05$ for the shoulder stretch and $0.10$ for the triceps
stretch), yet the final system reaches F1 $0.80$ and $0.94$ on them. This is because
a set contains clips from all four limbs, and propagation transfers the evidence of the
arm clips to the leg clips. The remaining errors concentrate in pairs of variants of
one exercise, such as plain and complex push-ups. These look the same from the head
and differ only in how the limbs move.

On the four hidden test participants, the configuration of Table~\ref{tab:prf} reaches
$0.871$ macro-F1 on the part of the test set that the challenge scores
publicly~\cite{wear2026challenge}.

\subsection{What component removal takes out}
\label{sec:removal}

\begin{figure*}[t]
\centering
\includegraphics[width=\textwidth]{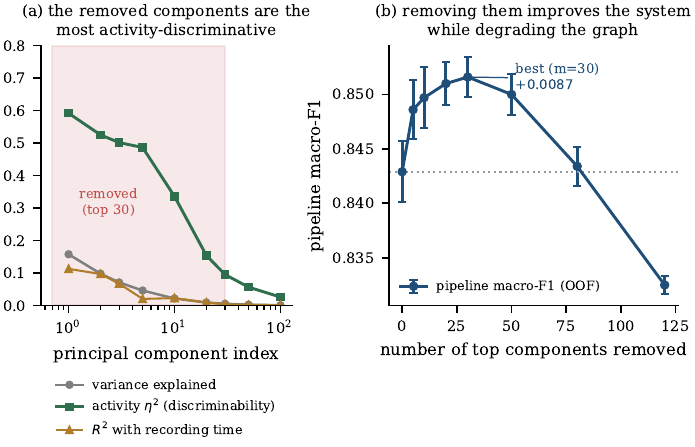}
\Description{Left: variance share, activity discriminability and time correlation of the leading principal components, with the top 30 shaded. Right: pipeline macro-F1 as a function of the number of removed components, peaking at 30.}
\caption{(a) The leading components contain both the most variance and by far the
most activity discriminability. (b) Removing them nevertheless raises macro-F1, with an interior optimum at 30, while neighbourhood activity purity falls
monotonically.}
\label{fig:pcdrop}
\end{figure*}

\paragraph{Removing the leading principal components raises macro-F1, although they are the most discriminative.}
Removing the top $m$ principal components of the per-participant video embedding
before building the graph raises macro-F1 (Figure~\ref{fig:pcdrop}b), with an
interior optimum at $m=30$ ($0.8429 \pm 0.0028$ at $m=0$, $0.8516 \pm 0.0018$ at
$30$, $0.8434$ at $80$ and $0.8325$ at $120$). The removed components are not a
nuisance subspace. Within a participant, components 1--30 contain $77.2\%$ of the
variance and a mean activity $\eta^2$ of $0.266$, against $0.047$ for components
31--100 (Figure~\ref{fig:pcdrop}a). They also contain the activity information that
transfers between participants. Retrieving a clip's neighbours from \emph{other}
participants gives label purity $0.408$ on the raw features and $0.198$ after
removing 100 components (chance $0.161$). Neighbourhood purity within a participant
also falls ($0.732$ at $m=0$, $0.633$ at $m=30$), so standard graph-quality measures
predict the opposite of what the system does. The removal operates on directions
rather than on coordinates. Deleting 400 of the 768 channels changes macro-F1 by
only $+0.0027$. Truncated PCA whitening, which keeps the top 128 or 256 directions and equalises their
variance, performs comparably to removal ($0.8527$ and $0.8535$, within our threshold of
$0.005$). The same truncation without whitening does not ($0.8415$ and $0.8425$). This suggests
that the weighting of directions in the similarities matters, but it does not separate removing
specific content from correcting anisotropy. The analyses below describe what the leading
directions contain (Appendix~\ref{app:falsify}, item~9).

\paragraph{The effect of removal is consistent across encoders.}
We encoded the public WEAR videos of the training participants with
CLIP~\cite{radford2021clip}, DINOv2~\cite{oquab2024dinov2} and
VC-1~\cite{majumdar2023vc1} (eight frames per clip, see Appendix~\ref{app:impl}).
We replaced only the features that the graph is built from and kept every
classifier fixed. Removing the top 30 components increases macro-F1 for all four encoders in both seeds (Table~\ref{tab:encgain}). The gain is $+0.007$ to $+0.015$
with the video-probability block in the graph and $+0.012$ to $+0.023$ when the
graph is built from the encoder's features only. For each encoder, $m=30$ is the
best setting tried. The improvement is largest for the encoder pretrained on egocentric video. The rest
of this section asks what the removed directions contain and which part of that
content the graph is better without.

\begin{table}[t]
\centering
\caption{Macro-F1 on WEAR when only the graph's features are replaced (20
participants, 61{,}404 clips, mean of 2 seeds). The classifiers are fixed, and the
per-clip prediction without a graph scores $0.644$. ``+prob.'': the graph also
contains the video-probability block. ``features only'': it does not.}
\label{tab:encgain}
\setlength{\tabcolsep}{4pt}
\begin{tabular}{lcccccc}
\toprule
 & \multicolumn{3}{c}{+prob.} & \multicolumn{3}{c}{features only} \\
\cmidrule(lr){2-4}\cmidrule(lr){5-7}
Encoder & $m=0$ & $m=30$ & $\Delta$ & $m=0$ & $m=30$ & $\Delta$ \\
\midrule
VideoMAEv2 & 0.8410 & 0.8480 & $+0.0070$ & 0.8339 & 0.8455 & $+0.0116$ \\
CLIP & 0.8342 & 0.8435 & $+0.0093$ & 0.8175 & 0.8351 & $+0.0176$ \\
DINOv2 & 0.8337 & 0.8429 & $+0.0092$ & 0.8165 & 0.8332 & $+0.0167$ \\
VC-1 & 0.8317 & 0.8462 & $+0.0145$ & 0.8151 & 0.8384 & $+0.0233$ \\
\bottomrule
\end{tabular}
\end{table}

\paragraph{The removed components are not limited to the view.}
If the gain came from discarding the view, subtracting the part of the embedding
that the view explains should reproduce it. It does the opposite
(Table~\ref{tab:removal}). Subtracting the component explained by the scene layout,
which holds half to two thirds of the variance, reduces macro-F1 by $0.006$ to $0.018$. Subtracting
what the visible body explains changes nothing, and neither does subtracting what the
camera tilt or the camera shake explains. We estimate the camera tilt from each frame
with GeoCalib~\cite{veicht2024geocalib}, which recovers the measured head tilt on
Ego-Exo4D with $R^2 = 0.75$. GeoCalib is least reliable on views without a horizon or
straight edges, such as close views of the ground or the sky, and such views are
frequent at the ends of the leading components (Figure~\ref{fig:pcends}). We measure
camera shake by phase correlation, which recovers the measured inter-frame head
rotation on Ego-Exo4D with $R^2 = 0.89$. Removing only components 1--2 yields a sixth
to a quarter of the improvement from removing 1--30. The view lets the graph group clips
recorded at the same place and moment, and the graph needs it. The same view also
encodes a take's task across recordings on Ego-Exo4D. There, removal makes the
transfer of labels between takes less accurate ($0.744 \to 0.724$, $0.743 \to 0.661$, $0.740 \to 0.713$ at $m=30$).

\begin{table}[t]
\centering
\caption{Change in macro-F1 on WEAR when a component is subtracted from the video
features before the graph is built (features-only graph, 2 seeds). Components are
the part of the embedding explained, within a participant, by the named quantity.
$^\ast$Uses ground-truth labels and is diagnostic only.}
\label{tab:removal}
\setlength{\tabcolsep}{3pt}
\begin{tabular}{lcccc}
\toprule
Subtracted & VideoMAEv2 & CLIP & DINOv2 & VC-1 \\
\midrule
top 30 components & $+0.012$ & $+0.018$ & $+0.017$ & $+0.023$ \\
components 1--2 only & $+0.002$ & $+0.004$ & $+0.003$ & $+0.004$ \\
\midrule
scene layout & $-0.013$ & $-0.009$ & $-0.006$ & $-0.018$ \\
visible body & $-0.002$ & $-0.001$ & $-0.001$ & $-0.002$ \\
camera tilt & $+0.000$ & $+0.001$ & $-0.000$ & $-0.001$ \\
camera shake & $+0.001$ & $+0.000$ & $+0.001$ & $+0.000$ \\
\midrule
what the four limbs predict & $+0.010$ & $+0.011$ & $+0.010$ & $+0.010$ \\
activity means$^\ast$ & $+0.013$ & $+0.022$ & $+0.030$ & $+0.017$ \\
what the classifier $P$ predicts & $-0.018$ & $-0.019$ & $-0.014$ & $-0.025$ \\
\bottomrule
\end{tabular}
\end{table}

\paragraph{The removed components contain a part of the embedding that the limbs predict.}
The gain is reproduced by subtracting from each clip's embedding the out-of-fold
prediction of that embedding from the four limbs' inertial features. With the
features-only graph, this recovers $41$--$83\%$ of the gain of removal (least for
VC-1, whose embedding the limbs predict least). A control that subtracts the same
predictions permuted among clips is worse than no subtraction. The limbs cannot
predict the view. None of scene layout, visible body, camera tilt or camera shake
explains the limb-predicted component ($R^2 \le
0.05$). The limbs do, however,
predict the activity. The per-participant activity means of the
embedding hold a third of its variance, and $84$--$86\%$ of that part is contained
in the limb-predicted component (VideoMAEv2, CLIP, DINOv2). Subtracting the activity
means themselves is a diagnostic that uses the labels. For the three generic
encoders it matches or exceeds removing the top 30 components ($+0.013$, $+0.022$,
$+0.030$ against $+0.012$, $+0.018$, $+0.017$), and for VC-1 it yields three quarters
of that improvement ($+0.017$ against $+0.023$). The harm is not a correlation between the
classifier's errors and the video. We shuffled the classifier's predictions among
the clips of the same participant and class, which preserves its confusions and removes
any link to the video. This leaves the gain in place and even enlarges it. On the
other hand, subtracting what the classifier's own predictions explain reduces macro-F1,
possibly because it imports the classifier's errors into the video.

\paragraph{Which clips removal corrects.}
Appendix~\ref{app:falsify} shows which clips removal corrects. These are mostly
\nul{} clips captured by the activity next to them and static stretches captured by
\nul{}. The appendix also compares removal with subtracting the limb-predicted
component, which a test clip carrying one limb cannot compute. These analyses describe
what removal changes. They do not establish that removing this content is what causes the
gain, since whitening, which keeps the leading directions, gives a similar gain.

\subsection{Role separation versus fusion}
\label{sec:roles}

\paragraph{The graph source sets the sign.}
Table~\ref{tab:2x2} crosses the classifier modality with the graph modality,
measured at the configuration before component removal. It reports the change
due to propagation against the same classifier without it. The video graph
improves both classifiers, and the inertial graph degrades both. Whether the two
modalities agree does not predict the sign. Propagation improves the same-modality pair (video, video)
and degrades the cross-modality pair (video classifier, inertial graph).
Agreement affects only the magnitude (the video graph raises the inertial classifier by $+0.167$ and the video classifier by $+0.037$). This is consistent with
the two representations being close to orthogonal. Regressing inertial features on
the video embedding explains only $0.21$--$0.25$ of their variance, and the mean
canonical correlation between the two spaces is $0.101$.

\begin{table}[t]
\centering
\caption{Change in macro-F1 due to propagation against the same classifier without
it (three seeds, configuration before component removal). Modality agreement is
annotated. It does not predict the sign, but the graph source does.}
\label{tab:2x2}
\setlength{\tabcolsep}{5pt}
\begin{tabular}{lcc}
\toprule
 & inertial graph & video graph \\
\midrule
inertial classifier & $-0.060$ \emph{(same)}  & $\best{+0.167}$ \emph{(cross)} \\
video classifier    & $-0.016$ \emph{(cross)} & $+0.037$ \emph{(same)} \\
\bottomrule
\end{tabular}
\end{table}

\paragraph{The graph source has a larger effect than the propagation rule.}
Table~\ref{tab:roles} repeats the comparison with LAME~\cite{boudiaf2022lame},
which adjusts classifier outputs as $z_i \propto p_i \odot \exp(\beta \sum_j W_{ij}
z_j)$. In its original setting, with the graph built from the classifier's own
(inertial) features, LAME changes macro-F1 by $+0.001$. The same LAME on the video graph yields $+0.118$, and our propagation rule on the video graph by $+0.158$. Changing the graph source changes macro-F1 by an order of magnitude more than changing
between these two propagation rules.

\begin{table}[t]
\centering
\caption{Graph source and fusion (single stage unless stated, two seeds). Base:
the inertial classifier alone, $0.6485$. Gains here and in Table~\ref{tab:2x2} are
measured in reduced configurations and are not directly comparable with those of
Table~\ref{tab:prf}.}
\label{tab:roles}
\setlength{\tabcolsep}{3pt}
\begin{tabular}{llc}
\toprule
Graph & Rule / fusion & $F_1$ \\
\midrule
inertial (classifier's features) & LAME, best $\beta$ & 0.6493 \\
inertial (classifier's features) & ours & 0.6025 \\
video & LAME, best $\beta$ & 0.7668 \\
video & ours & \best{0.8065} \\
\midrule
\multicolumn{3}{l}{\emph{full system, first-stage graph varied}} \\
video only & --- & \best{0.8523} \\
per-clip weights (WNN-style) & fusion & 0.7856 \\
uniform half & fusion & 0.7627 \\
inertial only & --- & 0.6836 \\
\bottomrule
\end{tabular}
\end{table}

\paragraph{Fusion performs worse in every form we tried.}
These variants were tuned less than the proposed system (Section~\ref{sec:setup}), so
the comparison shows which designs worked here rather than that fusion is worse in general. In the full system, a classifier consuming both modalities is weaker than the
inertial one on its own ($0.608$ vs.\ $0.648$) and far weaker after propagation
($0.663$ vs.\ $0.852$). Blending the two gives the strongest single classifier
($0.739$), yet after propagation it too performs worse ($0.783$). One possible explanation is that fusing at the classifier puts the video's view of the activity into every clip's
prediction, where the graph can no longer correct it. In a two-stage version of the
system without video pooling, adding inertial $k$-NN edges to the video graph
degrades it monotonically ($0.810$ with the video graph only, $0.709$ with 20
inertial edges per clip, one seed).
Learning view weights by minimising the smoothness of the propagated distribution
under each view's Laplacian~\cite{nie2016autoweighted} converges to an almost even
split ($0.518$ video, $0.482$ inertial). In a single-stage version of the system,
these weights reduce macro-F1 by $0.067$ relative to the video graph only. The criterion itself
differs by only $11\%$ between the views, while macro-F1 differs by $0.19$. We also
tried per-clip weights in the style of weighted nearest
neighbours~\cite{hao2021wnn}, in which each clip weights the two similarities by
how well each modality's neighbours explain its own profile. These weights become
sharply bimodal (mean video weight $0.61$) and reduce macro-F1 by $0.067$ (Table~\ref{tab:roles}).

\paragraph{Propagation already performs the sensor fusion.}
A teacher that uses all four sensors is a much stronger clip classifier than the
single-sensor inertial classifier ($0.762$ vs.\ $0.648$ out-of-fold). Yet
distilling it into one of the inertial networks changes the system by $0.0001$, and
using it directly as the base is worse than the weaker student ($0.812$ vs.\
$0.852$). The reason is that propagation already integrates
the four placements. Only $0.253$ of a clip's video neighbours wear its limb, which is
the chance rate, and restricting the graph to same-limb clips reduces macro-F1 by $0.118$. Further
analysis of the base classifier's contribution is in Appendix~\ref{app:base}.

\subsection{What the stages do}
\label{sec:stages}

Table~\ref{tab:stages} tests the reading of Section~\ref{sec:m-prop} by holding the
sharpening and pooling schedule fixed and changing only what the graph is rebuilt
from. More propagation on the same graph does not reproduce the gain of the stages.
Repeating three rounds without rebuilding ($-0.032$) or running fifteen iterations
in one round ($-0.022$) is no better than a single stage. Rebuilding improves the result, and the
source of the rebuild determines the sign. With the refined $Q$, the fraction of edges
joining clips of different true activities halves, from $0.194$ to $0.087$.
With the unrefined inertial prediction $P$, it rises to $0.245$ and macro-F1
decreases by $0.114$. With a permuted $Q$, it collapses. Rebuilt from the true labels, the
graph has no such edges and the system reaches $0.929$, close to the level of the set
oracle. The stages are therefore an alternating estimation of labels and affinity
that works once the labels have been refined.

\begin{table}[t]
\centering
\caption{What the stages do: macro-F1 (two seeds) and the fraction of edges joining
clips of different true activities at stages 1/2/3. The sharpening and pooling
schedule is identical in every row.}
\label{tab:stages}
\setlength{\tabcolsep}{3pt}
\begin{tabular}{lcc}
\toprule
Graph between stages & $F_1$ & Cross-activity edges \\
\midrule
rebuilt from refined $Q$ (ours) & \best{0.8523} & 0.194 / 0.087 / 0.087 \\
single stage & 0.8303 & 0.194 \\
not rebuilt, three rounds & 0.8204 & 0.194 / 0.194 / 0.194 \\
not rebuilt, 15 iterations & 0.8299 & 0.194 \\
rebuilt from base $P$ & 0.7379 & 0.194 / 0.245 / 0.245 \\
rebuilt from permuted $Q$ & 0.7396 & 0.194 / 0.703 / 0.604 \\
\midrule
rebuilt from true labels (oracle) & 0.9289 & 0.194 / 0.000 / 0.000 \\
\bottomrule
\end{tabular}
\end{table}

\subsection{Robustness and sensitivity}
\label{sec:unseen}

\paragraph{Participant identity.}
The graph compares clips of one participant only. Participant-specific cues such
as place, clothing and light therefore serve to recognise a set rather than act as a
nuisance. A graph pooled across the four participants of a test-sized group, with
no penalty on cross-participant pairs, forms only $5.1\%$ cross-participant edges,
of which $37\%$ join different activities. It leaves the score unchanged
($+0.0007$). The per-clip classifiers are applied across participants. We measured their reliance
on participant identity by training the same LightGBM on other sets of the
\emph{same} participant instead of on other participants. This raises macro-F1 by only $+0.0014$ on inertial features ($0.6005$ vs.\ $0.5991$) and $+0.009$ on video
features ($0.4556$ vs.\ $0.4465$), with a per-participant range of $-0.18$ to
$+0.27$ for video. The video, whose reliance varies more, enters only through the
within-participant graph and the pooled probabilities.

\paragraph{Where unseen participants fail.}
Final macro-F1 ranges from $0.64$ to $0.93$ across the 22 participants. It tracks
the inertial classifier's score ($r = 0.88$) and the set-oracle ceiling ($r =
0.87$) but not the purity of the video graph ($r = 0.12$). The lowest score is largely a labelling error.
Public work~\cite{goodpjw2026links} matched the training files to the per-session table of the WEAR
supplementary material~\cite{bock2024wear} and found that this participant's last session
($1{,}419$\,s) is labelled \nul{} throughout although nine activities were recorded in it. The same holds for
the last $112$\,s of another participant. Our system labels $62\%$ of these $1{,}531$ clips as activities,
which the released labels count as errors. Without them, this participant scores $0.78$, and the range
becomes $0.72$ to $0.93$. Of the other three hardest participants, two fail at the \nul{} boundary, where
gentle exercises and stretches are predicted as \nul, and the third in exercise variants. Deciding \nul separately has no benefit. The \nul class is detected worst from the video only (AUC $0.849$), better
from the inertial classifier ($0.878$) and best from the refined joint distribution
($0.942$). Every two-level rule we tried scores below the joint 19-way decision
($-0.011$ to $-0.095$).

\paragraph{Clips available in parts.}
The method treats a participant's clips jointly. If they become available in
parts, as when a stream is processed every so often, the same system run separately
on consecutive parts of the session retains most of its gain. It scores $0.853$, $0.857$ and
$0.844$ with halves, quarters and tenths of the session, against $0.852$ with all
clips and $0.648$ for the inertial classifier. Each 95\% interval of the
difference from $0.852$ includes zero. The parts are contiguous in time, so this
models a deployment in which clips arrive in order. It does not model the shuffled
test set. Only the graph and the propagation are restricted to the part. The class prior of the
decision rule is still averaged over all clips, which such a deployment would not have.

\paragraph{Decision rule and video rows.}
Two definitions in our evaluation could have been made differently. Averaging the class
prior $\pi_c$ of the decision rule over each participant's clips, rather than over all
clips being labelled, raises macro-F1 by $+0.019$ to $+0.020$ for every propagated
configuration (three seeds; $0.8516 \to 0.8707$ for the final system) and by $+0.002$ for
the inertial classifier. The gain from propagation therefore grows ($+0.182 \to +0.199$), while
the gains between the propagated configurations are nearly unchanged. We did not evaluate it
on the hidden test set, where a per-participant class balancing that gained in
cross-validation lost (Appendix~\ref{app:negative}). Building the graph from the video rows
that the test set supplies, rather than from the rows of Appendix~\ref{app:impl}, changes
the propagated configurations by $+0.000$ to $+0.002$ ($0.8518$ for the final system).

\paragraph{Sensitivity to hyperparameters.}
No one-at-a-time change of the hyperparameters improves macro-F1 on both seeds
(Appendix~\ref{sec:sens}).

\section{Discussion}
\label{sec:discussion}

\subsection{What the graph means in the domain}
\label{sec:disc-domain}
Read in terms of the activity, the graph connects the moments of a participant's
session in which the head camera saw the same thing: the same place, the same
posture and the head pointing the same way. A set of one exercise lasts about half a
minute (median $32$\,s) and holds all three roughly constant, so the graph is a stand-in for set
membership. The graph is useful because a clip's video neighbours often carry sensors on other limbs. An ankle clip during
an arm exercise contains almost no activity information, but its video neighbours include the same
set seen from the other limbs (the neighbours of the clip in
Figure~\ref{fig:teaser} are spread 6/9/9/6 over the four placements). Propagation
therefore performs at test time the four-sensor fusion that the protocol makes unavailable to each clip. When we rebuild the graph from refined predictions, ``the same scene'' narrows to
``the same scene and apparently the same exercise''. This separates exercises
performed in similar postures. The video cannot supply the exercise
variant. Plain and complex push-ups look the same from the head and differ in how
the limbs move. Video thus supplies the neighbourhood structure, and inertial sensing supplies the class
evidence.

Section~\ref{sec:encode} makes the same division inside the video embedding. Its
leading directions hold two things. One is the view, which lets the graph group the
moments of a session. The other is the appearance of the activity, which the limb sensors can predict. From the head, the appearance of an exercise is the appearance of its place
and posture. The pauses around a set and the stillness of a static stretch
share this appearance. The clips that removal corrects fit this picture: they are mostly
pauses captured by the activity next to them and static stretches captured by \nul{}.
Subtracting the limb-predicted part also reproduces much of the gain. Whitening the
leading directions without discarding them does as well, however, so we cannot say whether the gain comes from
removing this content or from changing how much the leading directions weigh in the
similarities. The analysis of what the embedding contains stands on its own. That its
removal causes the gain remains a hypothesis.

\subsection{Implications for the evaluation protocol}
\label{sec:disc-bench}

\paragraph{Shuffling does not remove adjacency.}
The test clips are consecutive, non-overlapping tiles. When two consecutive
clips carry the same limb (a quarter of pairs), the accelerometer signal
continues across their boundary, and the successor of such a clip can often be
found (Appendix~\ref{app:challenge}). The publicly released solution of
K.~Rudra builds on this signal~\cite{rudra2026repo}. Added on top of our system, it contributes little: smoothing
the predictions along the recovered links improves cross-validation macro-F1 by only $+0.004$. Systems built
around it gain much more. A public pipeline that recovers the order and propagates along the links scores
$0.906$ on the public leaderboard, and an extension with learned links reaches $0.929$ (nested
cross-validation $0.930$)~\cite{goodpjw2026links}, against $0.871$ for our system (Section~\ref{sec:main}). The same work reports that replacing each link by a random clip of the same bout
scores higher ($0.958$) than the true successor ($0.929$), which agrees with our view that what helps is
finding clips of the same set, not the order. We do not use the links, because they are a property of
how the test set was cut rather than of activity recognition. Our results should be
read as what is achievable when order is genuinely absent. Leaving gaps
between test clips in future editions would close this channel at no cost to the task.

\paragraph{Protocol knowledge as a test-time constraint.}
WEAR has a fixed protocol, in which each participant performs all eighteen activities for
roughly 90--120\,s. In the training recordings an activity class occupies a median
of $94$ one-second clips (5--95\% range $78$--$126$), and $431$ of $432$
recording--class pairs are non-empty. Yet $31\%$ of the recording--class counts
predicted by our system fall outside that range. Following the count calibration
of the same solution~\cite{rudra2026repo}, we add the protocol as a
constraint. For each participant, we adjust per-class biases on the prior-corrected scores
until every activity class has between $lo$ and $hi$ predicted
clips. \nul absorbs the remainder, so its share stays free (it varies from
$0.14$ to $0.57$ with recording length). With the range set from the
\emph{other} training recordings, this increases macro-F1 by $+0.0160 \pm 0.0012$ (three seeds, all
positive). The largest gains are on lunges and push-ups. The constraint raises macro-F1 more for the four hardest participants than for the rest ($+0.057$ vs.\ $+0.019$), and it is most effective where
variants are confused. Confusing two variants pushes one count above the
range and the other below it. The constraint is close to class mass
normalisation~\cite{zhu2003harmonic}. It differs from per-participant Sinkhorn
balancing, which fixes every class including \nul and did not transfer
(Appendix~\ref{app:negative}). It uses no test labels but assumes that test
participants followed the protocol. This holds for WEAR by design and not for
free-living data.

\section{Limitations}
\label{sec:limitations}
All results come from one dataset. The design choices were made on the same
cross-validation that we report (Section~\ref{sec:setup}), so the scores are optimistic
by an amount we do not measure, and the hidden test set is the only independent
check. All scores use the labels as released. Besides the two unlabelled stretches of
Section~\ref{sec:unseen}, one session has its labels $10$\,s late~\cite{goodpjw2026links}.
Leaving out the former and moving the latter raises the final macro-F1 from $0.852$ to $0.865$ and the inertial
classifier from $0.648$ to $0.658$ (three seeds), so the gains are unchanged. We keep the released labels,
since the test labels presumably follow the same conventions. The video rows of our training clips are offset from the test rows by up to eight
frames. Aligning the graph's rows changes nothing, but we did not retrain the
classifiers on aligned rows (Appendix~\ref{app:impl}). Cross-validation covers 22 participants and the hidden
test four, and the variation between participants is large (Section~\ref{sec:unseen}). The approach makes two assumptions.
First, a participant's clips are available together, although a tenth of a session at a
time retains most of the gain (Section~\ref{sec:unseen}). Second, an activity set is visually
consistent from the head. Free-living data, in which people move between places
during an activity or do different activities in one place, may violate the second
assumption. The $2\times2$ design of Table~\ref{tab:2x2} confounds invariance with
discriminativeness, since the video graph is also the stronger representation in
absolute terms. The negative claim (modality agreement does not determine the
sign) is therefore established more firmly than any positive one. Our account of component
removal rests on measurements of the view made with off-the-shelf models
(segmentation, single-image calibration, phase correlation). We validated these on
Ego-Exo4D but not on WEAR. The component this account identifies cannot be computed at
test time, where each clip carries one limb. We have not shown that removing it is what
causes the gain: whitening the leading directions gives a similar gain, and the two
explanations are not separated (Section~\ref{sec:removal}). The Ego-Exo4D check covers four encoders, one of them pretrained on
egocentric video. However, it uses our own encoding,
a different camera, sports rather than workouts, six participants and one exocentric
camera per take. The encoder comparison on WEAR
excludes three recordings (one that matched its public video only weakly and two
whose public video we did not download). Whether removal
improves results on egocentric tasks other than activity recognition is untested.

\section{Conclusion}
\label{sec:conclusion}
The leading directions of egocentric video embeddings from pretrained encoders primarily reflect where the camera looks. We observe this across four encoders, including one pretrained
on egocentric video, and not for exocentric cameras. It reflects what a head camera
records: where, and in which posture, an activity takes place, and far less of the
movement that makes it that activity. For combining such video with
body-worn sensors, this suggests a division of roles, and our comparisons are
consistent with it. In this setting, the video is most useful for finding the clips of the same set, and the
inertial classifier for labelling them. The view makes the video a good grouping signal. Reducing the dominance of the leading
directions, by removing or whitening them, improves the graph further. Whether this works by
removing the appearance of the activity that comes with the view, or by re-weighting the
directions in the similarities, we have not separated.
Every fusion variant we tried, in one classifier or in the graph edges, performed worse
than the division of roles, although none was tuned as extensively as the proposed system. Re-estimating the
graph from refined labels makes a multi-stage schedule effective. The propagation that implements the division is not new. What we add is evidence on when it
works in this setting, and an analysis of what the removed components contain. The analysis also exposes two properties of the benchmark, recoverable
adjacency and a strong class-count prior, that future shuffled-clip evaluations should take
into account.

\paragraph{Code availability.}
The code that reproduces every table and figure will be made publicly available
under the MIT Licence.

\appendix

\section{Implementation Details}
\label{app:impl}

\paragraph{Classifiers.}
The 1D-CNNs use four convolutional blocks (kernel sizes $7, 5, 3, 3$ and
$64 \to 128$ channels, with stride 2 twice), mean and max pooling, a 32-dimensional placement
embedding and dropout $0.3$. The network on both modalities and the video-only
network take the 15 video rows of a clip. The video-only probabilities are the
geometric mean of two networks trained on clips cut at different offsets. All
classifiers are trained with the subject-wise folds of Section~\ref{sec:setup}, and
we use out-of-fold predictions for every analysis.

\paragraph{Video feature construction.}
Following the WEAR feature specification, features are extracted at 30\,fps from
frames resized to $224\times224$. Row $i$ is computed from the 16-frame input made of
frame $i$, the 8 preceding and the 7 following frames. A one-second clip covers
frames $[f_0, f_0+30)$, and the 15 rows supplied per test clip are those with
$i \in [f_0+8, f_0+22]$, so the video side spans the same second as the inertial
side. Our training and validation clips were cut differently, a choice made early in development:
the graph features, the network on both modalities and one video-only network use rows
$[f_0, f_0+14]$, and the other video-only network uses $[f_0+7, f_0+21]$. Rows before $f_0+8$
draw on up to eight frames ($0.27$\,s) before the clip. Building the validation graph from
the test-aligned rows changes the final macro-F1 by $+0.0002$ (Section~\ref{sec:unseen}). We
did not retrain the classifiers on the aligned rows. Consecutive rows share 15 of
16 frames, so the 15 rows are not independent observations. For Ego-Exo4D we
compute three 16-frame inputs per clip (centred on frames $f_0+8$, $f_0+15$ and $f_0+22$)
and average them. The CLIP (\texttt{openai/clip-vit-base-patch32}, projected image
embedding), DINOv2 (\texttt{facebook/dinov2-base}, pooled output) and VC-1
(\texttt{facebook/vc1-base}, class token after the final normalisation) embeddings
use the single frames $f_0+8$, $f_0+15$ and $f_0+22$, averaged in the same way. For
the WEAR encoder comparison (Table~\ref{tab:encgain}) we downloaded the public WEAR
videos of the training participants only. We encode the frames $f_0, f_0+2, \dots,
f_0+14$ of each clip, matching the rows that the VideoMAEv2 feature averages. Public
file numbers differ from the challenge's for four recordings. We matched every
video to its recording by cross-correlating the frame-to-frame change of its
embedding with that of the challenge features (all at zero lag, $r = 0.74$--$0.84$
against at most $0.16$ for the next candidate). We excluded one recording that matched
only weakly ($r = 0.40$) and the two whose video we did not download.

\paragraph{Measuring the view on WEAR\@.} For the analyses of
Section~\ref{sec:encode} we take one frame per clip, at the centre of the feature rows the graph averages.
\emph{Scene layout and visible body}: SegFormer-B2~\cite{xie2021segformer} trained on
ADE20K~\cite{zhou2017ade20k}, with class fractions in a $3\times3$ grid. The ``person'' pixels
are almost always the wearer's own hands and legs. The scene classes are those
with a mean fraction above $0.2\%$ (18 classes). \emph{Camera tilt}:
GeoCalib~\cite{veicht2024geocalib} with its distorted-camera model. On the Aria frames
of Ego-Exo4D, its gravity estimate predicts the measured head tilt with $R^2 = 0.75$
(cross-validated over takes). \emph{Camera shake}: phase correlation between
consecutive frames at 10\,fps ($160\times90$ grey). We summarise it per clip by the mean
absolute shift in $x$ and $y$, their standard deviations, the largest shift and the
mean frame difference. On Ego-Exo4D it predicts the measured inter-frame head
rotation with $R^2 = 0.89$. We subtract components by ridge regression of the
per-participant centred embedding on the named quantity within each participant.
\emph{Exocentric control}: the first static camera of each Ego-Exo4D take, encoded in
the same way as the egocentric video.

\paragraph{Hyperparameters.}
$k=30$ (mutual), $\alpha=0.75$, five iterations per stage, three stages,
$\lambda=4$, sharpening exponent $1.4$, $m=30$, $\gamma=0.5$, $g=0.2$, $\tau$ tuned
out-of-fold per configuration ($0.9$ for the final system). All were selected on subject-wise cross-validation. The one-at-a-time sensitivity is reported below. The transductive
stages take $0.6$\,s per participant in the median and at most $1.8$\,s ($1{,}900$--$6{,}600$ clips) on an
8-core desktop CPU, without a GPU. The point at which video enters
matters. Pooling after every stage ($0.8429$) is slightly above pooling once after the last
stage ($0.8411$) and no pooling ($0.8301$). The dose--response in $g$ is
interior-optimal, and a confidence-gated weight performs worse than the uniform one.

\paragraph{Sensitivity to hyperparameters.}
\label{sec:sens}
We varied one parameter at a time around the final configuration ($k \in
\{15,\dots,90\}$, zero to three rebuilds, $\alpha \in [0.6, 0.85]$, 3--12
iterations, sharpening exponent 1.0--2.0, $g \in [0.1, 0.4]$, $m \in [20, 50]$,
$\gamma \in \{0, 0.25, 1\}$). No setting improves on both seeds. The settings
$k \ge 60$, $\alpha \le 0.6$, no sharpening, $g \ge 0.4$ and no rebuilding each reduce macro-F1 by
$0.006$ to $0.022$. The schedule is at a local optimum.

\section{Additional Results}
\label{app:perclass}

\begin{table}[t]
\centering
\caption{Per-class F1 of the inertial-only classifier, the video-only classifier
and the final system (seed 42), with the inertial classifier's recall split by arm
and leg sensors. Bold marks the stronger single modality where the gap exceeds
$0.05$.}
\label{tab:perclass}
\small
\setlength{\tabcolsep}{3pt}
\begin{tabular}{lccccc}
\toprule
 & \multicolumn{3}{c}{F1} & \multicolumn{2}{c}{inertial recall} \\
\cmidrule(lr){2-4}\cmidrule(lr){5-6}
class & inertial & video & final & arm & leg \\
\midrule
stretching (hamstrings) & 0.57 & \best{0.73} & 0.84 & 0.49 & 0.71 \\
lunges (complex) & 0.51 & \best{0.66} & 0.81 & 0.65 & 0.36 \\
stretching (shoulders) & 0.31 & \best{0.39} & 0.80 & 0.47 & 0.05 \\
sit-ups & 0.57 & 0.61 & 0.86 & 0.58 & 0.60 \\
jogging (rotating arms) & 0.68 & 0.69 & 0.84 & 0.82 & 0.53 \\
sit-ups (complex) & 0.62 & 0.61 & 0.84 & 0.57 & 0.74 \\
stretching (triceps) & 0.49 & 0.47 & 0.94 & 0.76 & 0.10 \\
\nul & 0.76 & 0.73 & 0.88 & 0.79 & 0.76 \\
push-ups (complex) & \best{0.60} & 0.53 & 0.75 & 0.52 & 0.62 \\
burpees & \best{0.67} & 0.59 & 0.87 & 0.63 & 0.69 \\
lunges & \best{0.54} & 0.45 & 0.79 & 0.47 & 0.57 \\
stretching (lumbar rot.) & \best{0.78} & 0.69 & 0.92 & 0.81 & 0.84 \\
push-ups & \best{0.65} & 0.55 & 0.78 & 0.69 & 0.68 \\
bench-dips & \best{0.62} & 0.51 & 0.86 & 0.77 & 0.46 \\
jogging & \best{0.77} & 0.58 & 0.87 & 0.90 & 0.70 \\
stretching (lunging) & \best{0.58} & 0.34 & 0.86 & 0.49 & 0.61 \\
jogging (sidesteps) & \best{0.85} & 0.57 & 0.92 & 0.80 & 0.93 \\
jogging (butt-kicks) & \best{0.84} & 0.51 & 0.88 & 0.74 & 0.87 \\
jogging (skipping) & \best{0.85} & 0.43 & 0.92 & 0.80 & 0.91 \\
\midrule
macro-F1 & 0.645 & 0.560 & \best{0.854} & & \\
\bottomrule
\end{tabular}
\end{table}

\begin{figure*}[t]
\centering
\includegraphics[width=\textwidth]{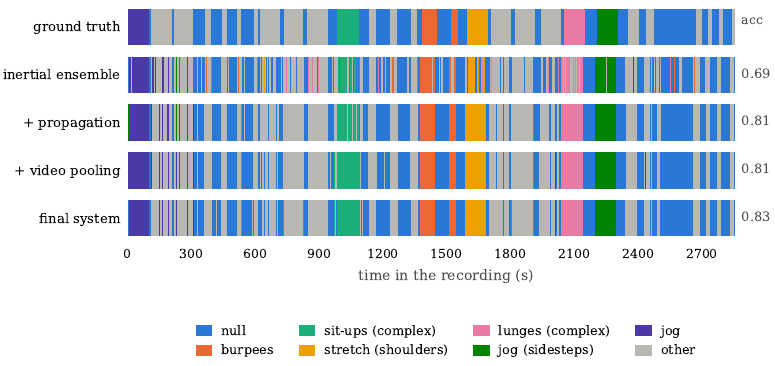}
\Description{Colour-coded label timelines for one recording: ground truth, the inertial ensemble, and three stages of the system, which become progressively closer to the ground truth.}
\caption{Ground truth and each stage for one participant (chosen as the one whose
gain from the last component is closest to the median). Temporal order is used for
this figure only, and the method never uses it.}
\label{fig:timeline}
\end{figure*}

\begin{figure*}[t]
\centering
\includegraphics[width=\textwidth]{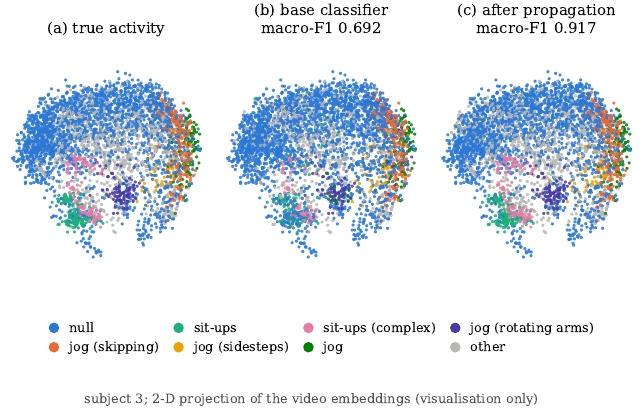}
\Description{Three two-dimensional projections of one participant's video embeddings coloured by true activity, base classifier prediction and refined prediction.}
\caption{Propagation on real data. Points are one participant's clips in a 2-D
projection of their video embeddings, and colour is the activity. (a) ground truth;
(b) the inertial classifier; (c) after propagation. macro-F1 on this participant
rises from $0.692$ to $0.917$.}
\label{fig:refine}
\end{figure*}

Table~\ref{tab:perclass} breaks the three predictors down by class. Video outperforms the
inertial classifier in only five classes. Taking the better modality per class
would raise macro-F1 only from $0.645$ to $0.668$. The gain to $0.854$ comes from
pooling across placements through the graph rather than from choosing a modality per
class. Figure~\ref{fig:timeline} shows the predictions over one participant's
recording, and Figure~\ref{fig:refine} shows the effect of propagation in the video
embedding.

\section{The Base Classifier's Contribution}
\label{app:base}

\begin{figure}[t]
\centering
\includegraphics[width=0.62\textwidth]{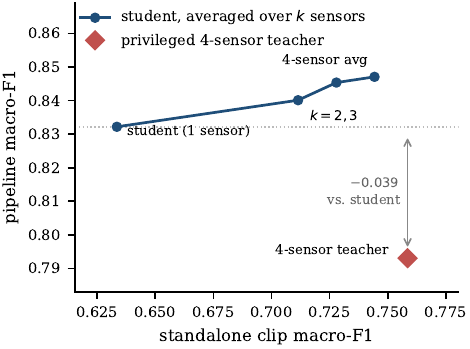}
\Description{Scatter plot of standalone clip macro-F1 against pipeline macro-F1 for students averaged over one to four sensors and for a privileged four-sensor teacher.}
\caption{At the base position, higher standalone accuracy yields little system accuracy.
Averaging the student over $k$ sensors of the same clip (blue, privileged)
traces a slope of about $0.13$. The four-sensor teacher (red) lies below this line.}
\label{fig:seat}
\end{figure}

\begin{figure*}[t]
\centering
\includegraphics[width=\textwidth]{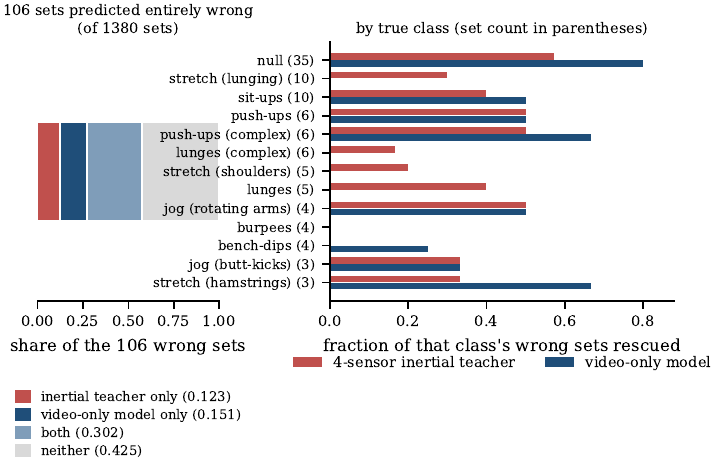}
\Description{Left: stacked bar of which model would rescue the sets predicted entirely wrong. Right: per-class fraction of wrong sets rescued by the four-sensor teacher and by the video-only model.}
\caption{Decomposition of the 106 sets predicted entirely wrong. Only $42.5\%$ are
missed by both a four-sensor inertial teacher and a video-only model, and $15.1\%$ are
recoverable by video alone. The split depends on the class, and video dominates on \nul.}
\label{fig:residual}
\end{figure*}

Averaging the single-sensor classifier's predictions over all four sensors of the
same clip (privileged, unavailable at test) raises its standalone accuracy by
$+0.109$ but the system by only $+0.014$ (Figure~\ref{fig:seat}). A $+0.01$ system
gain requires about $+0.077$ standalone. Biasing graph edges towards cross-placement pairs has no effect either (cross-placement neighbour rate $0.75
\to 0.98$, macro-F1 $0.8322 \to 0.8312$). The oracle averages four views of the
same second, whereas the graph averages different seconds of the same set. Of the 106
sets that an earlier configuration predicted entirely wrong, $42.5\%$ are missed by
both a four-sensor teacher and a video-only model, and $15.1\%$ are recoverable by
video alone (Figure~\ref{fig:residual}).

\section{Explanations Tested for the Effect of Component Removal}
\label{app:falsify}

\paragraph{Look-alike clips are captured by the wrong class.}
The clips that removal corrects are those whose video neighbourhood belonged to
another activity. Their neighbourhood purity in the raw graph is $0.42$--$0.54$,
against $0.75$--$0.80$ overall. Of their raw errors, $67$--$71\%$ (with the
top 30 removed, and $75$--$79\%$ with the activity means subtracted) are exactly the
class that surrounded them. The captures follow one pattern in all four encoders.
\nul{} clips are captured by the activity next to them (jogging, sidesteps, the
sit-up and push-up variants, burpees, bench-dips), and static stretches are captured by
\nul. Subtracting the activity means lifts the neighbourhood purity of these clips
from $0.42$--$0.46$ to $0.75$--$0.78$. The gain arrives through re-estimation. A
single stage achieves a fifth to two fifths of the improvement from three stages. Across classes, the gain
is predicted by how much re-estimation reduces the edges to other activities
(Spearman $\rho$ $0.75$--$0.96$ for removal, $0.64$--$0.81$ for subtraction). In
domain terms, a head camera records where and in which posture an activity is done,
and far less of the movement that defines the activity. The seconds before and after a set,
labelled \nul, are spent at the same spot in the same posture, and a static stretch
looks like standing still. Their appearance therefore matches the neighbouring activity,
while the limbs differ in whether they move. The appearance component pulls such
clips into the look-alike class, propagation overrides their inertial evidence,
and re-estimation fixes the error in place. Without it, the view still groups the
clips of a session, and the inertial classifier assigns the labels.

\paragraph{The limb-predicted component is unavailable at test time.}
For VideoMAEv2, with the video-probability block in the graph, subtracting the
limb prediction gives $0.8502$, against $0.8448$ on the raw features and $0.8523$ with
removal (Table~\ref{tab:partial}). Subtraction and removal are partly complementary
($0.8555$ together). Predicting the video better brings no further gain, since gradient
boosting has the highest $R^2$ and the smallest gain. Projecting out a fixed
subspace (the ten video directions most correlated with the limbs) has no
effect. Only subtracting a per-clip amount set by that clip's limbs improves the result.

A test clip carries one limb, and the account does not transfer to it. A
prediction from one limb improves the result when it is the \emph{same} limb for every clip
($0.8544$), but not when each clip uses the limb it wears ($0.8511$). Different
limbs explain different parts of the view, which distorts distances between
clips wearing different limbs. Centring per placement has no effect. The
missing limbs also cannot be completed well enough to restore consistency. A rank-120
factorisation reconstructs all four limbs' features almost perfectly ($R^2 =
0.96$). From one limb and the video graph, however, they are recovered only to $R^2 =
0.37$ by graph-regularised tensor completion, and to $0.41$ at best. The best
test-applicable variants change macro-F1 by $+0.0011$, which is within the seed-to-seed noise. The
account therefore explains the effect of removal in part, but it offers no way to exploit it
at test time.

Estimating the appearance component without the limbs does not raise the full system either (video-probability block and evidence pooling included, three seeds).
An appearance subspace learned from the other participants' labels (the span of
their activity means, or discriminant directions) improves macro-F1 over no removal ($+0.002$),
against $+0.008$ for removing the top 30 components. It brings no further improvement on top of that
removal. The appearance of an activity depends on where and facing which way a
participant performed it, and it does not transfer between participants. Estimating the
activity means within the test participant from the system's own predictions imports
their errors ($-0.001$ to $-0.003$ with soft and $-0.008$ to $-0.020$ with hard
assignments, relative to the current system). Even the ground-truth activity means raise macro-F1 by only $+0.0045$ over the current system, because the probability block and
component removal already absorb most of the harm.

\begin{table}[t]
\centering
\caption{Graph built on video features after subtracting a limb-predicted
component (two seeds). We denote by $m$ the number of leading components removed
afterwards. ``Test'' marks variants a test clip can supply.}
\label{tab:partial}
\setlength{\tabcolsep}{3pt}
\begin{tabular}{lccc}
\toprule
Subtracted component & $m=0$ & $m=30$ & Test \\
\midrule
none (raw features) & 0.8448 & 0.8523 & \checkmark \\
four limbs, MLP & 0.8502 & \best{0.8555} & \\
four limbs, linear & 0.8500 & 0.8553 & \\
four limbs, gradient boosting & 0.8491 & 0.8535 & \\
four limbs, MLP, permuted & 0.8372 & --- & \\
one limb, same for every clip & 0.8475 & 0.8544 & \\
one limb, the limb each clip wears & 0.8430 & 0.8511 & \checkmark \\
one limb, same-limb pairs only$^\dagger$ & --- & 0.8534 & \checkmark \\
one limb and video, factor model & 0.8457 & 0.8534 & \checkmark \\
top-10 CCA directions projected out & 0.8448 & 0.8519 & \checkmark \\
\bottomrule
\end{tabular}

\smallskip
{\footnotesize $^\dagger$ Pairs wearing the same limb are compared on the residual,
and all other pairs on the baseline features.}
\end{table}

We tested twenty-three explanations for why removing the leading video components
improves the graph. We rule out twenty. Two (items 8 and 12) fail with one limb
but hold in part with four, and Section~\ref{sec:removal} identifies what the four limbs capture.
One (item 9) we cannot rule out. Truncated whitening performs comparably to removal, and our
tests do not separate correcting anisotropy from removing specific content.

\begin{enumerate}\itemsep1pt
\item \textbf{Removing slow appearance drift.} Correlation of the components with
  recording time is weak ($R^2 \le 0.11$ for the first, $0.022$ over 1--30). Median
  neighbour time separation rises with $m$ ($58.8$\,s to $89.3$\,s), but purity is
  uniformly lower at every time lag.
\item \textbf{Improving neighbourhood purity.} Purity falls monotonically while the
  score peaks.
\item \textbf{Improving placement invariance.} The video graph is at chance for
  placement ($0.251$) before and after.
\item \textbf{Indirectly weakening propagation.} Lowering $\alpha$ at $m=0$ only
  reduces macro-F1.
\item \textbf{Indirectly reducing propagation depth.} Iterations are flat at $m=0$.
\item \textbf{Indirectly re-weighting the probability block of the graph
  features.} $\gamma$ is already at its optimum.
\item \textbf{Compensating a video--inertial offset.} Neighbour purity of the video
  graph against labels shifted by $s$ seconds peaks at $s=0$ before and after
  removal, with the same profile (Table~\ref{tab:shift}). This tests the data as a whole. It does
  not detect an offset in a single session, such as the $10$\,s label delay of Section~\ref{sec:limitations}.
\item \textbf{Reducing redundancy with the base classifier.} With predictions and a
  linear single-limb map, the candidate disappears. With a non-linear map from four
  limbs, it returns (Section~\ref{sec:removal}).
\item \textbf{Correcting anisotropy} (not ruled out). Centred features are strongly
  anisotropic, as for word vectors~\cite{mu2018allbut}. The first component holds $15.8\%$
  of the variance within a participant, the top 30 hold $77.2\%$, and the effective
  dimension is $19.5$ of $768$. Projecting onto the top 128 or 256 components and equalising their variance (truncated
  whitening, which discards the remaining components) gives $0.8527$ and $0.8535$, against
  $0.8516$ for removal and $0.8429$ for neither (three seeds). Projecting onto the same components
  without whitening gives $0.8415$ and $0.8425$, so the gain comes from equalising the variance, not from
  the truncation. A near-zero mean pairwise cosine ($0.0015$) does not guarantee isotropy.
\item \textbf{A coarse/fine division of labour.} $80.3\%$ of base errors cross
  coarse groups, and the leading components also contain within-group signal.
\item \textbf{Reducing hubness.} Hubs~\cite{radovanovic2010hubs} are more accurate
  than average, and explicit local scaling does not reproduce the gain.
\item \textbf{Residualising against the inertial features.} From one limb, linear
  residualisation is worth $-0.0007$ to $+0.0004$. From four limbs, it recovers a large part of the gain (Section~\ref{sec:removal}).
\item \textbf{Visibility of the wearer's body} (external labels from a pose
  detector). The features do not predict it across recordings, and the fitted
  directions are not enriched in the top-30 subspace ($0.025$--$0.042$ against
  $0.039$ for a random direction).
\item \textbf{Camera egomotion} (external labels from phase correlation). Its
  direction occupies $0.0392$ of the top-30 subspace, against $0.0391$ at random.
\item \textbf{Posture confusions shared with the classifier.} The correlation
  between the graph's confusions and the classifier's rises with removal ($0.636 \to
  0.800$), which is the opposite sign.
\item \textbf{Separating different sets of the same activity.} The same-set fraction
  among same-label neighbours is flat ($0.646$--$0.652$).
\item \textbf{A contaminated subspace estimate.} Estimating the subspace from the
  $70\%$ least-moving clips reduces macro-F1 by $0.0022$.
\item \textbf{Removing head orientation itself.} Projecting out the directions that
  measured head tilt spans on Ego-Exo4D, transferred to WEAR, changes macro-F1 by $+0.0003$ to
  $+0.0011$, as much as random directions. Subtracting the part explained by the
  camera tilt estimated on WEAR has no effect (Section~\ref{sec:removal}).
\item \textbf{Discarding the view.} Subtracting what the scene layout explains reduces macro-F1.
  Subtracting what the visible body or the camera shake explains changes nothing
  (Table~\ref{tab:removal}).
\item \textbf{Errors of the classifier that follow the video.} Shuffling the
  classifier's predictions among clips of the same participant and class retains
  the gain and enlarges it.
\item \textbf{Letting the label block steer re-estimation.} Raising $\lambda$ without
  removal lowers macro-F1.
\item \textbf{Purity dominated by \nul.} Class-averaged purity without \nul{} also
  falls with removal ($0.769 \to 0.673$).
\item \textbf{Scattering concentrated wrong neighbours.} Removal lowers the share of
  the largest wrong class among wrong neighbours, but across classes this share is
  negatively related to the gain.
\end{enumerate}

\begin{table}[t]
\centering
\caption{Neighbour purity of the video graph against labels shifted by $s$ seconds.}
\label{tab:shift}
\setlength{\tabcolsep}{4pt}
\begin{tabular}{lccccc}
\toprule
 & $s{=}-2$ & $s{=}-1$ & $s{=}0$ & $s{=}+1$ & $s{=}+2$ \\
\midrule
$m=0$  & 0.719 & 0.732 & \best{0.732} & 0.719 & 0.703 \\
$m=30$ & 0.621 & 0.632 & \best{0.633} & 0.622 & 0.608 \\
\bottomrule
\end{tabular}
\end{table}

\section{Frames at the Ends of the Leading Components}
\label{app:pcends}
\begin{figure}[t]
\centering
\includegraphics[width=\textwidth]{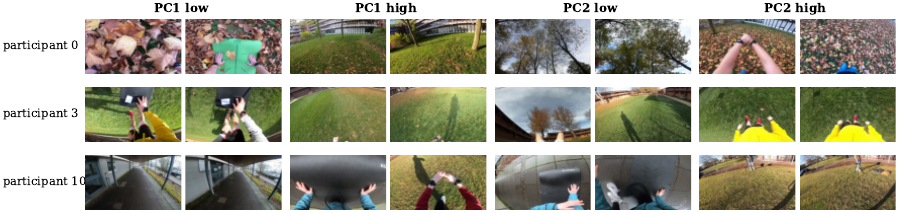}
\Description{Grid of camera frames for three participants: two frames at the low and high end of the first and second principal components of each recording's video embedding.}
\caption{Frames at the two ends of the first and second principal components of the
video embedding (VideoMAEv2, one recording per participant). Frames from
WEAR~\cite{bock2024wear} (CC BY-NC-SA 4.0, \url{https://creativecommons.org/licenses/by-nc-sa/4.0/}), downscaled.}
\label{fig:pcends}
\end{figure}

Figure~\ref{fig:pcends} extends Figure~\ref{fig:teaser}a to three participants. For
each recording, we centre the embedding and compute its first two principal
components. We show two frames at each end, at least 60\,s apart. At every end, the
frames share what the camera looks at, either the ground or the wearer's own body close
up, or the scene ahead or above. For participant 10, the first component also
separates two places (a covered walkway and a field).

\section{Other Negative Results}
\label{app:negative}
\begin{itemize}\itemsep1pt
\item \textbf{Labelled anchors.} Adding a vote from each test clip's top-30 video
  neighbours among labelled training clips reduces macro-F1 at every dose ($-0.012$ at
  $h=0.1$ to $-0.074$ at $0.35$), although the vote by itself scores $0.449$.
\item \textbf{Subject-adaptive self-training} on confident pseudo-labels: $0.832 \to
  0.70$--$0.75$.
\item \textbf{Per-class decision weights} (19 parameters): $+0.025$ in
  cross-validation, $-0.006$ on the hidden test set.
\item \textbf{Per-participant Sinkhorn balancing} with a uniform non-\nul target:
  $+0.017$ in cross-validation and $-0.038$ on the hidden test set. With a per-participant
  \nul target, there is no loss and no gain.
\item \textbf{Metric learning} on the video features (within-recording
  contrastive), hard clustering, confidence-weighted propagation and per-recording
  video standardisation: equal or worse.
\item \textbf{Temporal order recovery from video}: successor top-1 $\le 0.35$ by
  three methods.
\item \textbf{Cascaded propagation of a stacked classifier's output} and
  \textbf{graph ensembles}: about $+0.002$, below the decidable threshold.
\item \textbf{Mirror-canonicalised inertial features}, mapping left limbs onto the
  right-limb frame by sign flips chosen on training data: $+0.005$ for the
  inertial classifier on its own, and $+0.0002$ to $+0.0006$ in the full system (three
  seeds). Propagation largely absorbs improvements of the base classifier.
\end{itemize}

\section{Adjacency of Shuffled Clips}
\label{app:challenge}
When two consecutive clips wear the same limb, the last sample of one and the
first of the next are $20$\,ms apart. On training participants cut the same way, the
gap between them (median $0.052$\,g) is the size of one within-clip step
($0.047$\,g), against $1.25$\,g for an arbitrary same-limb pair. Consider the clips whose
successor wears the same limb. For $0.681$ of them, ranking same-limb candidates by the
extrapolated gap finds the true successor first, and adding video continuity raises
this to $0.750$. Among the most confident fifth, $99.2\%$ are correct. The signal is
weaker than these figures suggest, because three quarters of successors wear a
different limb, where only video continuity is available (top-1 $0.26$). It is also
weaker because near-static clips match by chance (best-candidate gap: median $0.023$\,g for
true successors, $0.065$\,g otherwise). A learned link scorer with a linear
assignment links half the clips with label agreement $0.95$. The smoothing of
Section~\ref{sec:disc-bench} is measured along these links.
The adjacency was measured on training participants and on the gap distribution of the test
set, and it is used in no reported result.

\end{document}